\pdfoutput=1
\documentclass[11pt]{article}
\usepackage{emnlp2021}
\usepackage[tikz]{bclogo}
\usepackage{times}
\usepackage{latexsym}

\usepackage[T1]{fontenc}
\usepackage[utf8]{inputenc}

\usepackage{lipsum}

\usepackage[english]{babel}

\usepackage{array, makecell}
\usepackage{amsmath,amsthm}
\usepackage{multirow}
\usepackage{microtype}
\usepackage{booktabs}
\usepackage{lipsum}  
\usepackage{longtable}
\usepackage{enumitem}
\usepackage{subcaption}
\usepackage{adjustbox}
\usepackage{tikz}

\usepackage[noend]{algorithmic}
\usepackage{algorithm}

\usepackage{inconsolata}
\usepackage{listings}

\colorlet{punct}{red!60!black}
\definecolor{background}{HTML}{EEEEEE}
\definecolor{delim}{RGB}{20,105,176}
\colorlet{numb}{magenta!60!black}

\lstdefinelanguage{json}{
    basicstyle=\scriptsize\ttfamily,
    numbers=left,
    numberstyle=\scriptsize,
    stepnumber=1,
    numbersep=8pt,
    showstringspaces=false,
    breaklines=true,
    frame=lines,
    backgroundcolor=\color{background},
    literate=
     *{0}{{{\color{numb}0}}}{1}
      {1}{{{\color{numb}1}}}{1}
      {2}{{{\color{numb}2}}}{1}
      {3}{{{\color{numb}3}}}{1}
      {4}{{{\color{numb}4}}}{1}
      {5}{{{\color{numb}5}}}{1}
      {6}{{{\color{numb}6}}}{1}
      {7}{{{\color{numb}7}}}{1}
      {8}{{{\color{numb}8}}}{1}
      {9}{{{\color{numb}9}}}{1}
      {:}{{{\color{punct}{:}}}}{1}
      {,}{{{\color{punct}{,}}}}{1}
      {\{}{{{\color{delim}{\{}}}}{1}
      {\}}{{{\color{delim}{\}}}}}{1}
      {[}{{{\color{delim}{[}}}}{1}
      {]}{{{\color{delim}{]}}}}{1},
}

\usepackage{microtype}

\newcommand{\benchmarkname}[1]{NLP Few-shot Gym}
\newcommand{\challengename}[1]{\textsc{CrossFit}}

\newcommand\crossfitemoji{\raisebox{-2pt}{\includegraphics[width=0.9em]{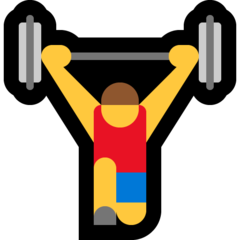}}}

\title{\vspace*{-0.5in}
{{\small \hfill \textit{in Proc. of EMNLP 2021}}\\
\vspace*{.25in}}
\textsc{CrossFit}~\crossfitemoji{}: A Few-shot Learning Challenge for \\ Cross-task Generalization in NLP}

\author{Qinyuan Ye \quad Bill Yuchen Lin \quad Xiang Ren \\
  University of Southern California \\
  \texttt{\{qinyuany, yuchen.lin, xiangren\}@usc.edu} 
}

\usepackage[textsize=footnotesize]{todonotes}

\definecolor{msftBlue}{RGB}{0,164,239}
\definecolor{msftGreen}{RGB}{127,186,0}
\definecolor{msftYello}{RGB}{255,185,0}
\definecolor{msftBlack}{RGB}{0,0,0}
\newcommand{\finding}[1]{
	\begin{bclogo}[couleur= msftBlack!05, epBord= 1, arrondi=0.1, logo=\bclampe,marge= 2, ombre=true, blur, couleurBord=msftBlack!10, tailleOndu=3, sousTitre ={\em #1}]{} 
	\end{bclogo}
}

\date{}

\begin{document}
\maketitle
\begin{abstract}

Humans can learn a new language task efficiently with only few examples, by leveraging their knowledge obtained when learning prior tasks. In this paper, we explore whether and how such \textit{cross-task generalization} ability can be acquired, and further applied to build better \textit{few-shot learners} across diverse NLP tasks.
We introduce \challengename{}~\crossfitemoji{}, a problem setup for studying cross-task generalization ability, which standardizes seen/unseen task partitions, data access during different learning stages, and the evaluation protocols.
To instantiate different seen/unseen task partitions in \challengename{} and facilitate in-depth analysis, we present the \benchmarkname{}, a repository of 160 diverse few-shot NLP tasks created from open-access NLP datasets and converted to a unified text-to-text format.
Our analysis reveals that the few-shot learning ability on unseen tasks can be improved via an upstream learning stage using a set of seen tasks. 
We also observe that the selection of upstream learning tasks can significantly influence few-shot performance on unseen tasks, asking further analysis on task similarity and transferability.\footnote{Our code is at  \url{https://github.com/INK-USC/CrossFit}.}

\end{abstract}

\section{Introduction}
\begin{figure}
    \centering
    \includegraphics[width=0.5\textwidth]{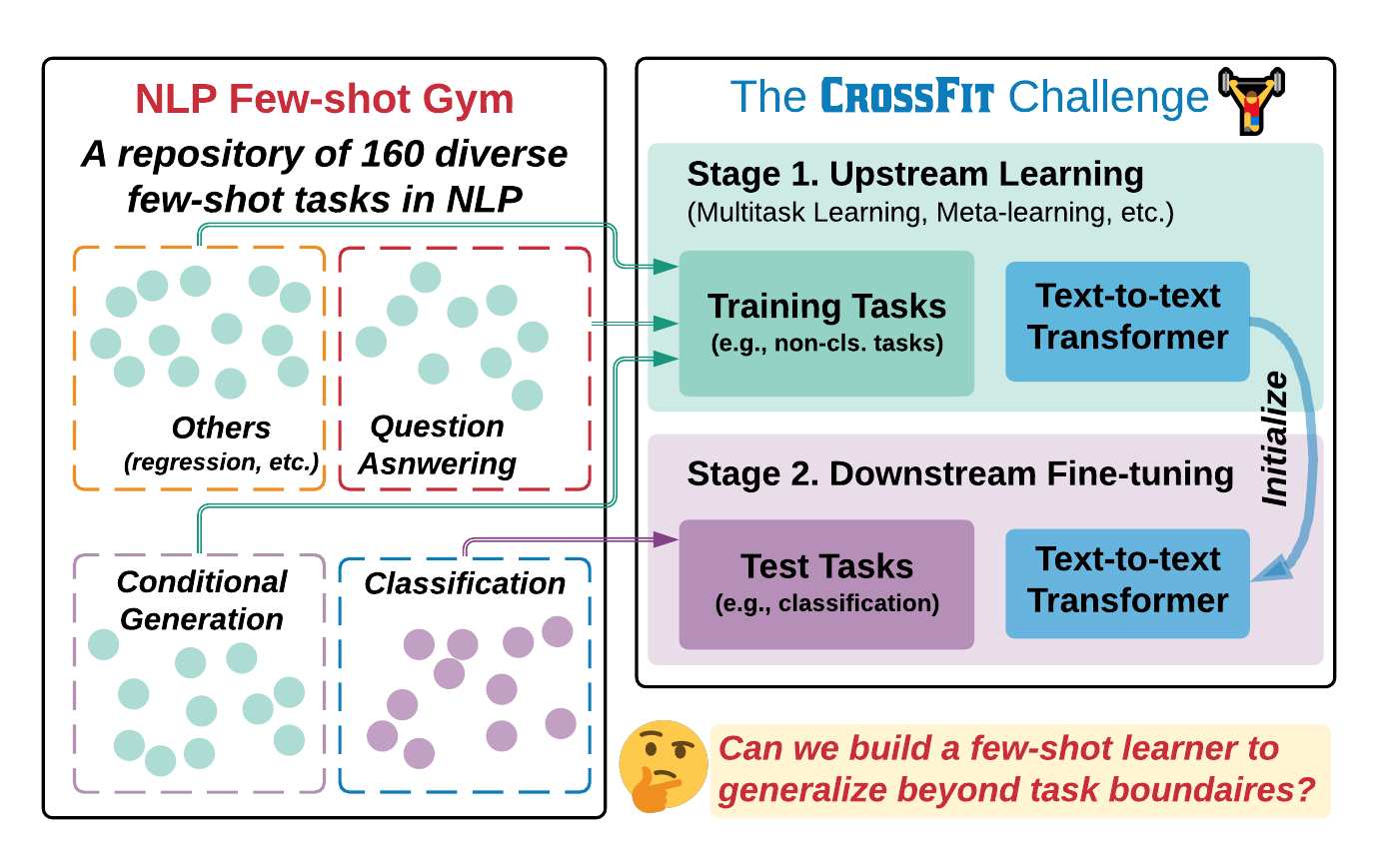}
    \caption{We present the \challengename{} Challenge to study cross-task generalization in a diverse task distribution. To support this problem setting, we introduce the \benchmarkname{}, a repository of 160 diverse few-shot, text-to-text tasks in NLP.}
    \label{fig:intro}
\end{figure}

Pre-trained language models fine-tuned with abundant task-specific data have become the predominant recipe for state-of-the-art results in NLP. 
However, these approaches are heavily dependent on large-scale labeled datasets that are expensive to create, and the resulting models still generalize poorly to out-of-distribution inputs created with small, harmless perturbations \cite{ribeiro-etal-2020-beyond}.
In retrospect, researchers have advocated for building more human-like, general linguistic intelligence that can ``reuse previously acquired knowledge about a language and adapt to a new task quickly'' \cite{Yogatama2019LearningAE, linzen-2020-accelerate}. 

Existing work has approached this problem via better few-shot fine-tuning, by re-formulating target tasks into cloze questions that resembles the pre-training objective \cite{schick2020exploiting,schick2020small}, generating prompts and using demonstrations \cite{Gao2020MakingPL}. Such progress primarily focus on improving \textit{instance-level generalization}, \textit{i.e.}, how to better generalize from few labeled instances to make predictions about new instances, \textit{within the scope of one individual task}. From a broader perspective, human-like learning ability also benefits from \textit{task-level generalization}, or \textit{cross-task generalization}, \textit{i.e.}, how to learn a new task efficiently given experiences of learning previous tasks.


Such ability has been widely studied in computer vision and robotics community \cite{pmlr-v100-yu20a, Triantafillou2020MetaDataset}, but is relatively under-explored in NLP. 
\citet{pruksachatkun-etal-2020-intermediate} and \citet{vu-etal-2020-exploring} study transferability between \textit{one} intermediate task and a given target task, while it's possible to further improve performance with \textit{multiple} intermediate tasks.
\citet{han-etal-2018-fewrel} and \citet{bansal-etal-2020-learning} focus on cross-task generalization within the scope of classification tasks, whereas humans can generalize across different task formats (classification, multiple choice, generation, etc.), goals (question answering, fact checking, etc.) and domains (biomedical, social media, etc.). 




Towards developing general linguistic intelligence,
we present \challengename{}, a few-shot learning challenge to acquire, evaluate and analyze cross-task generalization in a realistic setting, with standardized training pipeline, data access and evaluation protocol. 
The \challengename{} challenge requires a model to first learn from a set of seen tasks in an upstream learning stage, and then perform few-shot learning on a set of unseen tasks, as illustrated in Fig.~\ref{fig:intro}. 
In accompany, we introduce the \benchmarkname{}, a repository of 160 few-shot NLP tasks gathered from open-access resources, covering a wide range of capabilities and goals, and formulated into a unified text-to-text format. 
To analyze the capability and limitation of existing approaches to the \challengename{} challenge, 
we design eight specific seen/unseen task partitions.

With the \challengename{} Challenge and the \benchmarkname{}, we aim to investigate the following research questions:
\begin{itemize}[leftmargin=*,nosep]
\itemsep0em 
    \item \textbf{Q1.} Can we teach cross-task generalization ability to pre-trained models with existing methods? 
    \item \textbf{Q2.} During upstream learning, is it better to be ``well-rounded'' (learning from diverse tasks) or be ``specialized and targeted'' (learning from tasks in the same category with unseen tasks)? 
    \item \textbf{Q3.} Does it help if we have more labelled data for seen tasks during upstream learning?
\end{itemize}

To address the above questions, we empirically analyze the performance of multi-task learning and three meta-learning algorithms (MAML \cite{pmlr-v70-finn17a}, first-order MAML and Reptile \cite{Nichol2018OnFM}).
We observe that these approaches can indeed lead to better few-shot performance on unseen tasks. 
Interestingly, simple multi-task learning outperforms existing meta-learning methods in many cases, encouraging future research on identifying the reasons and developing improved meta-learning methods.
For Q2, we observe that performance of individual unseen tasks varies with different selection of seen tasks, calling for more thorough investigation of the relationship between task similarity and transferability.
As for Q3, we find that enlarging the size of upstream data does not necessitate better cross-task generalization abilities.
We envision cross-task generalization to be an integral component towards general linguistic intelligence, and we hope \challengename{} serves as a useful testbed for driving related progress.

\section{Related Work}
\paragraph{Few-shot Fine-tuning.}\label{sec:fsl}
Few-shot learning refers to teaching models a new task with a small number of annotated examples. 
Large-scale pre-trained language models (e.g., BERT~\cite{devlin-etal-2019-bert})
have demonstrated great ability to learn new tasks efficiently via \textit{fine-tuning} \cite{zhang2021revisiting}.
\citet{schick2020exploiting, schick2020small} proposed \textit{pattern-exploiting training} (PET), 
which formulates text classification and NLI tasks into cloze questions (or ``prompts'') that resemble masked language modeling.
PET can be further improved by generating prompts \textit{automatically} and incorporating demonstrations into the input \cite{Gao2020MakingPL}; and by densifying the supervision signal with label conditioning \cite{Tam2021ImprovingAS}.
While successful, in these approaches the downstream tasks are learned in isolation. Our work aims to boost few-shot learning ability on unseen tasks via acquiring cross-task generalization ability from diverse seen tasks.

\paragraph{Meta-learning in NLP.} Recent works have explored meta-learning methods for relation classification \cite{han-etal-2018-fewrel, gao-etal-2019-fewrel}, general text classification \cite{dou-etal-2019-investigating, bansal-etal-2020-learning, bansal-etal-2020-self}, low-resource machine translation \cite{gu-etal-2018-meta}, cross-lingual NLI/QA \cite{nooralahzadeh-etal-2020-zero}. In general, these works apply meta-learning algorithms to a set of sub-tasks; however the sub-tasks are either \textit{synthetic} (\textit{e.g.}, classifying a new set of five relations is a new sub-task) or drawn from a rather \textit{narrow} distribution (\textit{e.g.}, QA in one language is a sub-task). 
In our work, we explore a more realistic setting~--~learning from a set of NLP tasks with \textit{diverse} goals: classification, question answering, conditional generation, etc.
This setting is attracting attention in NLP community rapidly and is also explored in very recent work \cite{Zhong2021MetatuningLM,mishra2021natural, bragg2021flex, Wei2021FinetunedLM}.

\paragraph{Unifying NLP Task Formats.} 
Researchers have explored unifying the formats of different tasks, in order to better enable knowledge transfer, \textit{e.g.}, DecaNLP \cite{McCann2018TheNL}, UFO-Entail \cite{yin-etal-2020-universal} and EFL \cite{wang2021entailment}. 
Following T5 \cite{2020t5}, we adopt a unified text-to-text format that subsumes all text-based tasks of interest.
Related to our work, UnifiedQA \cite{khashabi-etal-2020-unifiedqa} examines the feasibility of training a general cross-format QA model with multi-task learning.
Our work extends from these ideas, and we significantly enlarge the task repository to 160 to broaden the coverage, in hopes to build a general-purpose few-shot learner.
\section{The \challengename{} Challenge}

In this section, we present the \challengename{} Challenge, a problem setting for acquiring and evaluating cross-task generalization.
Ideally, a strong \challengename{} system can capture cross-task generalization ability from a set of seen tasks and thus adapts to new unseen tasks efficiently.



\subsection{Preliminaries}
\label{ssec:preliminaries}
The meaning of ``task'' is overloaded: ``tasks'' can be categorized at different granularity (\textit{e.g.}, text classification vs. QA,  yes/no QA vs. machine reading comprehension), and from different aspects (\textit{e.g.}, domain, label space).
Herein we take a general formulation by defining a ``task'' with its training and testing examples. 
We define a task $T$ as a tuple of $(\mathcal{D}_{train}, \mathcal{D}_{dev}, \mathcal{D}_{test})$.
Each set $\mathcal{D}$ is a set of annotated examples $\{(x_i, y_i)\}$ in text-to-text format. 
In few-shot setting, the size of $\mathcal{D}_{train}$ and $\mathcal{D}_{dev}$ are required to be small (\textit{e.g.}, 16 example per class for classification tasks). 

Existing work mostly focuses on improving \textit{instance-level} generalization for individual task by using \textit{task-specific} templates.
Performance on individual tasks is used as the measure of success.
For the \challengename{} Challenge, we aim to acquire \textit{cross-task generalization} and build better \textit{general-purpose} few-shot learners, which calls for a different problem setting with distinct training procedure and evaluation protocol.


\subsection{Problem Setting}
\label{ssec:formulation}


\paragraph{Tasks and Data.}

To acquire and evaluate cross-task generalization, we first gather a large repository of few-shot tasks $\mathcal{T}$, and partition them into three non-overlapping sets $\mathcal{T}_{train}$, $\mathcal{T}_{dev}$, $\mathcal{T}_{test}$.
In hopes to examine the capability and limitation of an approach in different settings, and to answer our research questions, we design multiple task partitions with different focuses. 
Details of the repository and partitions, or as we name them, the \benchmarkname{}, are deferred to \S\ref{sec:forest}.



\paragraph{Learning Stages.}

A \challengename{} method may learn from $\mathcal{T}_{train}$ and perform necessary tuning with $\mathcal{T}_{dev}$ in the upstream learning stage; it is then evaluated with few-shot tasks in $\mathcal{T}_{test}$: 
\begin{itemize}[leftmargin=*, nosep]
\itemsep0em 
    \item \textbf{Upstream learning stage.}
    Here, the algorithm has access to the $\mathcal{D}_{train}$ and $\mathcal{D}_{dev}$ for each training task in $\mathcal{T}_{train}$, while $\mathcal{D}_{test}$ is unavailable. The algorithm also has access to all data in $\mathcal{T}_{dev}$, but for validation purpose only (\textit{i.e.}, it is not allowed to use $\mathcal{T}_{dev}$ to update model weights).
    \item \textbf{Few-shot learning stage.}
    In this stage, $\mathcal{T}_{test}$ became available. Models resulting from the upstream learning stage are required to learn from $\mathcal{D}_{train}$ via a particular few-shot learning method (\textit{e.g.}, direct fine-tuning). 
    The final few-shot learning performance is evaluated on $\mathcal{D}_{test}$. \footnote{For clarification, the performance on the $\mathcal{D}_{dev}$ of a task in $\mathcal{T}_{dev}$ or $\mathcal{T}_{test}$ will be used for tuning hyper-parameters during fine-tuning. The overall performance on $\mathcal{T}_{dev}$ is used for tuning tuning hyper-parameters during upstream learning.}  
\end{itemize}

\vspace{-0.1cm}
\paragraph{Evaluation Metric.}
Evaluating the performance of a model on a diverse collection of NLP tasks is inherently challenging, as different tasks use different metrics. 
It is thus not reasonable to simply aggregate performance of classification tasks (\textit{e.g.}, accuracy, F1) and generation tasks (\textit{e.g.}, ROUGE, BLEU) by taking the average.

To address this problem, 
we first narrow down to a collection of 7 evaluation metrics: classification F1, accuracy, QA F1, exact match (EM), Rogue-L, Matthew correlation, and Pearson correlation, which cover all tasks in our experiments.
Then, we define \textit{Average Relative Gain} (\textbf{ARG}), a metric that computes relative performance changes before and after the \textit{upstream learning stage} for each test task, and finally take the average across all test tasks. 

For example, suppose we have $\mathcal{T}_{test}=\{T_A, T_B\}$.
If an upstream learning algorithm helps improve the few-shot learning performance from $50\%$ F1 score to $70\%$ on task $T_A$ (\textit{i.e.}, a $40\%$ relative improvement), 
and from $40\%$ accuracy to $30\%$ on task $T_B$ (\textit{i.e.}, $-25\%$ relative improvement), 
the final ARG on $\mathcal{T}_{test}$ would be computed as $\frac{40\%+(-25\%)}{2}=7.5\%$.

The ARG metric reflects the \textit{overall} performance gain on all tasks in $\mathcal{T}_{test}$, no matter what specific metrics each task uses. We use ARG for a high-level comparison, and we still analyze the performance for each task (\textit{e.g.}, absolute performance metrics, performance growth with ``more shots'', sensitivity to different selection of $\mathcal{T}_{train}$) in our in-depth analysis.

\section{\benchmarkname{}}

\label{sec:forest}

Towards learning to generalize across tasks in \challengename{} challenge, 
we need a resource that contains sufficient number of tasks, 
covering a wide range of NLP applications, and presented in a unified text-to-text format. 
Herein, we introduce the \benchmarkname{}, a repository of 160 few-shot tasks gathered from existing open-access datasets.




\subsection{Dataset Selection}
\label{ssec:selection}
We choose to use Huggingface Datasets\footnote{\url{https://huggingface.co/datasets}.
It is an extensible library that provides access to 626 open-access NLP datasets (as of Feb 25th, 2021) with a unified, open-source API. } \cite{Lhoest2021DatasetsAC} as the pool of our candidate tasks.
We filter these datasets on a case-by-case basis, mainly using the following criteria: (1) We focus on English monolingual datasets. (2) We exclude datasets that require information retrieval, as they require a separate retriever. (3) We exclude sequence labeling tasks (\textit{e.g.}, dependency parsing, NER), which are highly dependent on tokenization, and are hard to evaluate in text-to-text format. 
(4) We exclude datasets dealing with extremely long documents (\textit{e.g.}, a scientific paper) as input, as most pre-trained models cannot process such long input sequences.
We finalize our selection with 160 datasets which are detailed in Appendix \ref{sec:all_tasks_table}.



\subsection{A Unified Text-to-Text Format} 
\label{ssec:unifying}
We follow \citet{2020t5} and convert all of our datasets into a unified text-to-text format.
For example, the task of natural language inference (originally a sentence-pair classification problem) 
becomes:  \texttt{premise: <premise> hypothesis: <hypothesis>}, and the target sequence is either the word \texttt{entailment}, \texttt{contradiction} or \texttt{neutral}.
As for machine reading comprehension tasks, 
the input format is 
\texttt{question: <question> context: <context>} and the target sequence is the correct answer span. 
We also reference the format for QA tasks from UnifiedQA \cite{khashabi-etal-2020-unifiedqa}.

\subsection{Formulating Few-shot Tasks} 
\label{ssec:sampling}


We mainly follow the practice in \cite{Gao2020MakingPL} for few-shot sampling. 
For classification and regression tasks, we include 16 training examples \textit{per class} in $D_{train}$. For other types of tasks, we include 32 examples in  $D_{train}$. In conformity with real-world situations where labeled data are scarce, we assume a development set $D_{dev}$
which shares the same size with $D_{train}$.

We sample $\mathcal{D}_{train}$ and $\mathcal{D}_{dev}$ splits from each dataset's original train set with 5 different random seeds.
This helps us reduce variance during few-shot evaluation, and also enlarges the number of few-shot tasks used for learning. 
Consequently, the ``effective size'' of our \benchmarkname{} is $160\times5=800$, while we use the number $160$ throughout the paper to avoid possible confusion. 

We use the original development set for each dataset as $\mathcal{D}_{test}$, or withhold $20\%$ of the dataset when the official development split is not available. 
The held-out test examples are sampled \textit{once} before sampling $\mathcal{D}_{train}$ and $\mathcal{D}_{dev}$.

\begin{figure}
    \centering
    \includegraphics[width=0.5\textwidth]{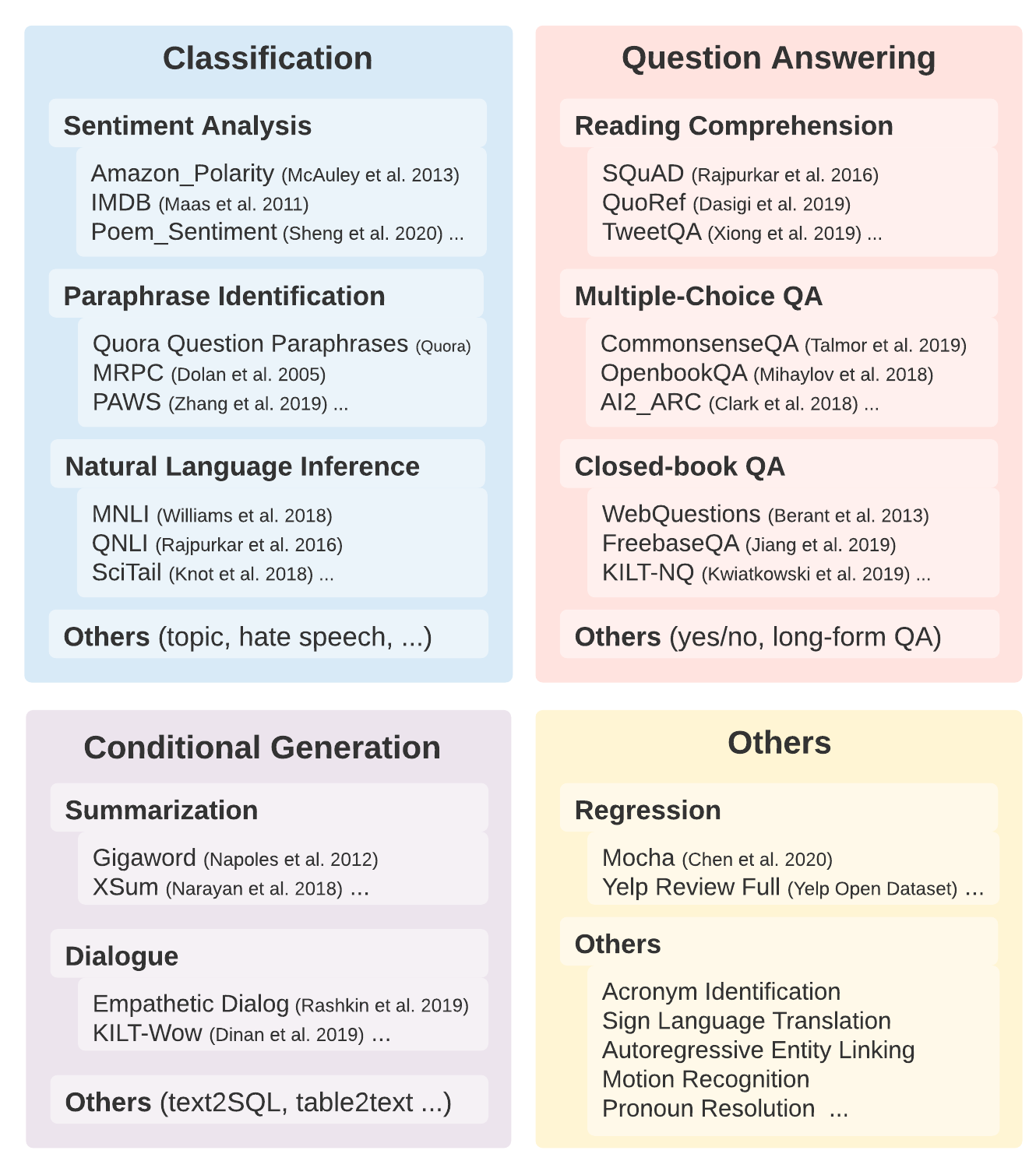}
    \caption{Task Ontology for the NLP Few-shot Gym. Full information is listed in Appendix~\ref{sec:all_tasks_table}.}
    \label{fig:ontology}
\end{figure}

\subsection{Task Ontology and Partitions}
\label{ssec:partitions}

As mentioned in \S\ref{ssec:formulation}, a \challengename{} method is expected to first acquire cross-task generalization on a set of $\mathcal{T}_{train}$ and evaluate such ability on $\mathcal{T}_{test}$. To comprehensively analyze to what extent a trained model can generalize, and how its behavior differs in different scenarios, we need to build different partitions of $(\mathcal{T}_{train}, \mathcal{T}_{dev}, \mathcal{T}_{test})$.

\begin{figure}
    \centering
    \includegraphics[width=0.48\textwidth]{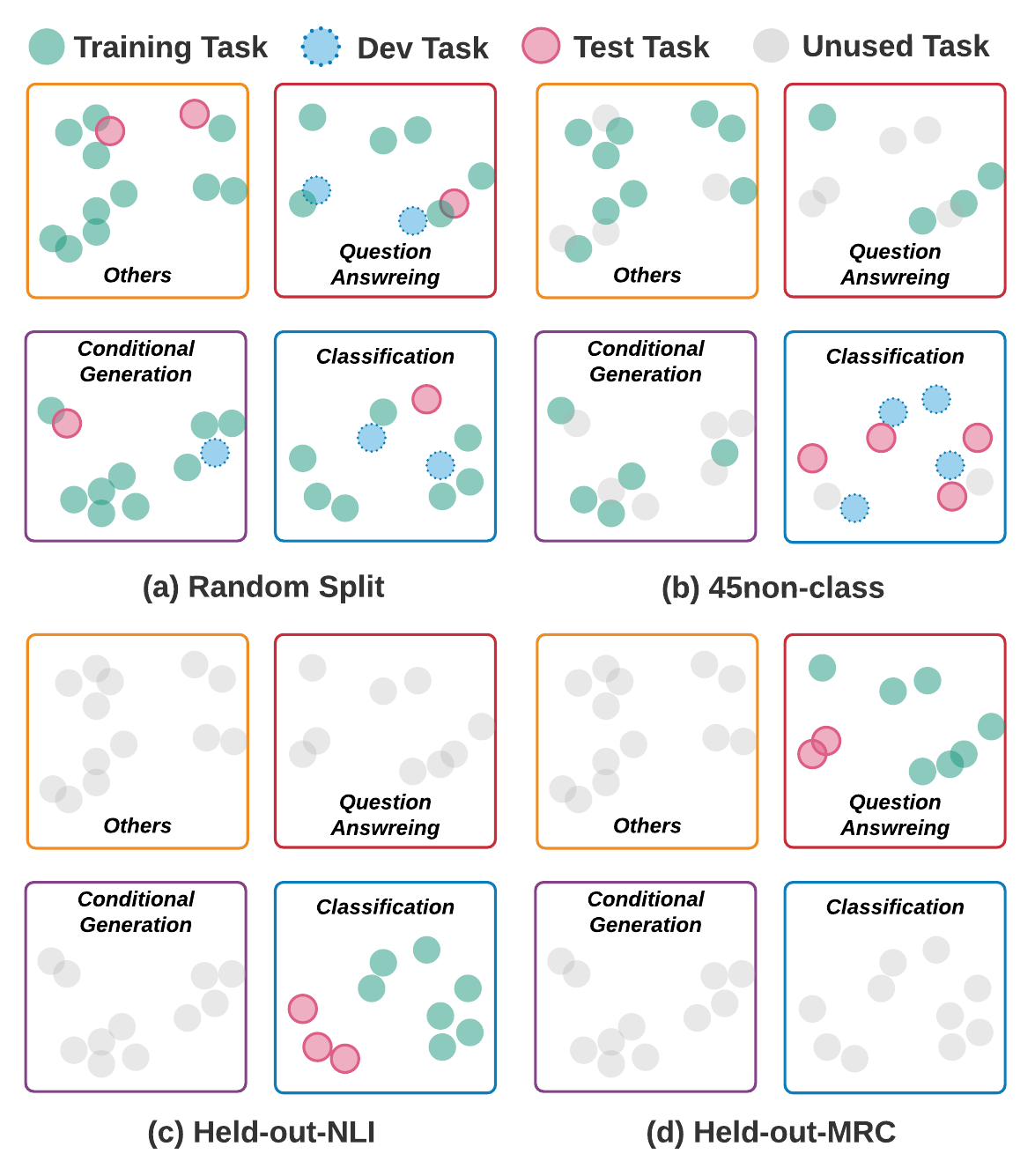}
    \caption{\textbf{Illustration for different task partitions.} We evaluate a \challengename{} approach on different task partitions to examine its generalization ability in different scenarios. Full details in Table~\ref{tab:splits}. The locations and distances in this figure are hypothetical and for illustrative purposes only.}
    \vspace{-0.5cm}
    \label{fig:splits}
\end{figure}

Towards this goal, we first manually classify the 160 tasks and form a \textbf{task ontology} with categories and sub-categories, as shown in Fig.~\ref{fig:ontology}. 
The first-level categories include classification, question answering, conditional generation, and others.\footnote{We later discuss the limitation of this design in \S\ref{sec:analysis}-Q2}
Further, we design eight different partitions of $(\mathcal{T}_{train}, \mathcal{T}_{dev}, \mathcal{T}_{test})$. We illustrate four partitions in Fig.~\ref{fig:splits} and provide more details in Table~\ref{tab:splits}. 

Our Partition 1 randomly split all 160 few-shot tasks into the three sets, where $|\mathcal{T}_{train}|=120$ and $|\mathcal{T}_{dev}|=|\mathcal{T}_{test}|=20$. The design of Partition 1 mimics the real-world language learning environment where the goal is to build a general-purpose few-shot learner, and a set of diverse tasks ($\mathcal{T}_{train}$) are used to train the learner.
Our Partition 2.1-2.3 withhold 10 classification tasks for development and 10 more for testing. The $\mathcal{T}_{train}$ is controlled to have either 100\% classification tasks, 100\% non-classification tasks, or half-and-half. These three partitions help us to understand the influence brought by different task distribution in $\mathcal{T}_{train}$. 
The remaining four partitions still focus on crossing task boundaries, but in a finer granularity: seen and unseen tasks are in the same category, but not the same sub-category. For example, Partition 3.1 has 57 non-NLI classification tasks as $\mathcal{T}_{train}$, and 8 NLI tasks as $\mathcal{T}_{test}$. These partitions help us to understand whether cross-task generalization in this finer granularity is easier for model to acquire.

\section{Methods to \challengename{}}
\label{sec:analysis}
\vspace{-0.1cm}

We mainly use BART-Base \cite{lewis-etal-2020-bart} as the text-to-text transformer for our analysis in the \challengename{} setup.
We leave confirmatory experiments with T5-v1.1-Base and BART-Large model in Appendix \ref{app:additional_exp}.

\paragraph{Direct Fine-tuning on Test Tasks.} 
This serves as the basic baseline method for the \challengename{} challenge, which does not make use of $\mathcal{T}_{train}$ or $\mathcal{T}_{dev}$, or go through the upstream learning stage.
For each task $T \in \mathcal{T}_{test}$, we directly fine-tune the text-to-text model with its $\mathcal{D}_{train}$, tune the hyper-parameters with $\mathcal{D}_{dev}$, and assess its performance with the test set $\mathcal{D}_{test}$.
We use the performance of direct fine-tuning as the base for computing \textbf{ARG} scores of other \challengename{} approaches. 
We expect a model trained with upstream learning would capture cross-task generalization ability and thus have better \textbf{ARG} scores.


\begin{table*}[]
\centering
\scalebox{0.65}{
\begin{tabular}{r|c||ccc|cccc|c}
\toprule
No. & Shorthand & $\mathcal{T}_{train}$ & $\mathcal{T}_{dev}$ & $\mathcal{T}_{test}$ & ARG(Multi) & ARG(MAML) & ARG(FoMAML) & ARG(Rept.) & Details \\\midrule
1    &   Random        &  120  &  20  &  20  &  $35.06\%$  & $28.50\%$  & $22.69\%$ & $25.90\%$ &    Fig.~\ref{fig:all}(a)  \\\midrule
2.1    & 45cls          &  45 cls.   &     {10 cls.}   &   {10 cls.}  &  $11.68\%$   &  $9.37\%$  & $10.28\%$ & $13.36\%$ & \multirow{3}{*}{Fig.~\ref{fig:classification-merge-test}}        \\
2.2    & 23cls+22non-cls          &  23 cls. + 22 non-cls.   &   {10 cls.}   &   {10 cls.}          &  $11.82\%$ & $9.69\%$ & $13.75\%$ & $14.34\%$&    \\
2.3    & 45non-cls         &  45 non-cls.   &    {10 cls.}   &   {10 cls.}     & $11.91\%$   & $9.33\%$ & $11.20\%$ & 
$14.14\%$&              \\\midrule
3.1    & Held-out-NLI  &   57 non-NLI cls.  &   /     &  8 NLI   &   $16.94\%$  & $12.30\%$ & $12.33\%$ & $14.46\%$ &  Fig.~\ref{fig:all}(b)\\
3.2    & Held-out-Para  &   61 non-Paraphrase cls.  &   /     &  4 Para. Iden.   &    $18.21\%$ & $17.90\%$ & $21.57\%$ & $19.72\%$ & Fig.~\ref{fig:all}(c) \\\midrule
4.1 & Held-out-MRC & 42 non-MRC QA & / & 9 MRC &$32.81\%$ & $27.28\%$ & $28.85\%$ & $28.85\%$ & Fig.~\ref{fig:all}(d) \\
4.2 & Held-out-MCQA & 29 non-MC QA & / & 22 MC QA & $12.20\%$ & $4.69\%$ &$6.73\%$ & $7.67\%$ & Fig.~\ref{fig:all}(e)\\\bottomrule
\end{tabular}
}
\caption{$(\mathcal{T}_{train}$,$\mathcal{T}_{dev}$,$\mathcal{T}_{test})$ partitions used in the study (full lists in Appendix~\ref{app:partitions}), and their ARG scores when upstream learning methods are applied. 
``cls.'' stands for ``classification'', ``Para. Iden.'' for ``paraphrase identification'', ``MRC'' for ``machine reading comprehension'' and ``MCQA'' for ``multiple-choice QA''.
}
\label{tab:splits}
\end{table*}



\paragraph{Multi-task Learning (MTL).} 
A straight-forward yet effective method is 
to combine the data\footnote{Both $\mathcal{D}_{train}$ and $\mathcal{D}_{dev}$ are used, as $\mathcal{D}_{dev}$ is used for gradient updates in meta-learning algorithm. 
We do so to make sure that the data access for the two methods is fair.} 
in the training
tasks to learn a multi-task model, before fine-tuning it on each test task.
Specifically, we gather source-target examples for all tasks in $\mathcal{T}_{train}$ and fine-tune the text-to-text model with these examples. 
Then we use the resulting checkpoint as initialization and perform the same procedure in ``direct fine-tuning'' for each test task in $\mathcal{T}_{test}$. 
The performance gain over the \textit{direct fine-tuning} is used for computing its overall ARG score.


\paragraph{Model-Agnostic Meta-learning (MAML).} 

Cross-task generalization ability, closely aligns with the concept of learning to learn. Hence, we use MAML \cite{pmlr-v70-finn17a}, a representative meta-learning approach during upstream learning.
The core concept of MAML is to learn a set of initialization weight, from which the model adapts fast to a new task within few gradient updates.
In MAML training, we iterate through tasks in $\mathcal{T}_{train}$ to update the model.
For each train task $(\mathcal{D}_{train}, \mathcal{D}_{dev})$, 
we first sample a support batch $\mathcal{B}_{support}$ from $\mathcal{D}_{train}$ and a query batch $\mathcal{B}_{query}$ from $\mathcal{D}_{dev}$. 
We use $f_{\theta}$ to denote the text-to-text model with parameters $\theta$. 
Using $\mathcal{B}_{support}$, we first compute the updated parameters $\theta'$ with gradient descent (\textit{i.e.}, the inner loop). Due to the large size of pre-trained text-to-text models, we use one gradient update in the inner loop, \textit{i.e.}, $\theta' = \theta - \alpha \nabla _{\theta} \mathcal{L}(f_{\theta}, \mathcal{B}_{support}).$
Then we apply the updated text-to-text model $f_{\theta'}$ to $\mathcal{B}_{query}$, and do one step of meta-optimization (\textit{i.e.}, the outer loop), with $\theta \leftarrow \theta - \beta\nabla_{\theta} \mathcal{L}(f_{\theta'}, \mathcal{B}_{query})$.

\paragraph{First-order MAML.} 
First-order MAML \cite{pmlr-v70-finn17a} avoids second-order optimization and improves training stability using the first-order approximation by differentiating with respect to the fast weights $\theta'$ instead of the original parameters $\theta$ for the gradient $\nabla_{\theta} \mathcal{L}(f_{\theta'}, \mathcal{B}_{query})$, \textit{i.e.}, $\theta \leftarrow \theta - \beta\nabla_{\theta'} \mathcal{L}(f_{\theta'}, \mathcal{B}_{query}).$

\paragraph{Reptile.} Reptile \cite{Nichol2018OnFM} is another memory-efficient, first-order meta-learning algorithm that first makes multiple gradient updates in the inner loop, then directly uses $\theta'-\theta$ to approximate $\nabla_{\theta} \mathcal{L}(f_{\theta'}, \mathcal{B}_{query})$, i.e., $\theta \leftarrow \theta + \beta (\theta' - \theta)$.


\section{Empirical Analysis}
\label{sec:analysis}



\begin{figure*}[t]
    \centering
    \includegraphics[width=1\textwidth]{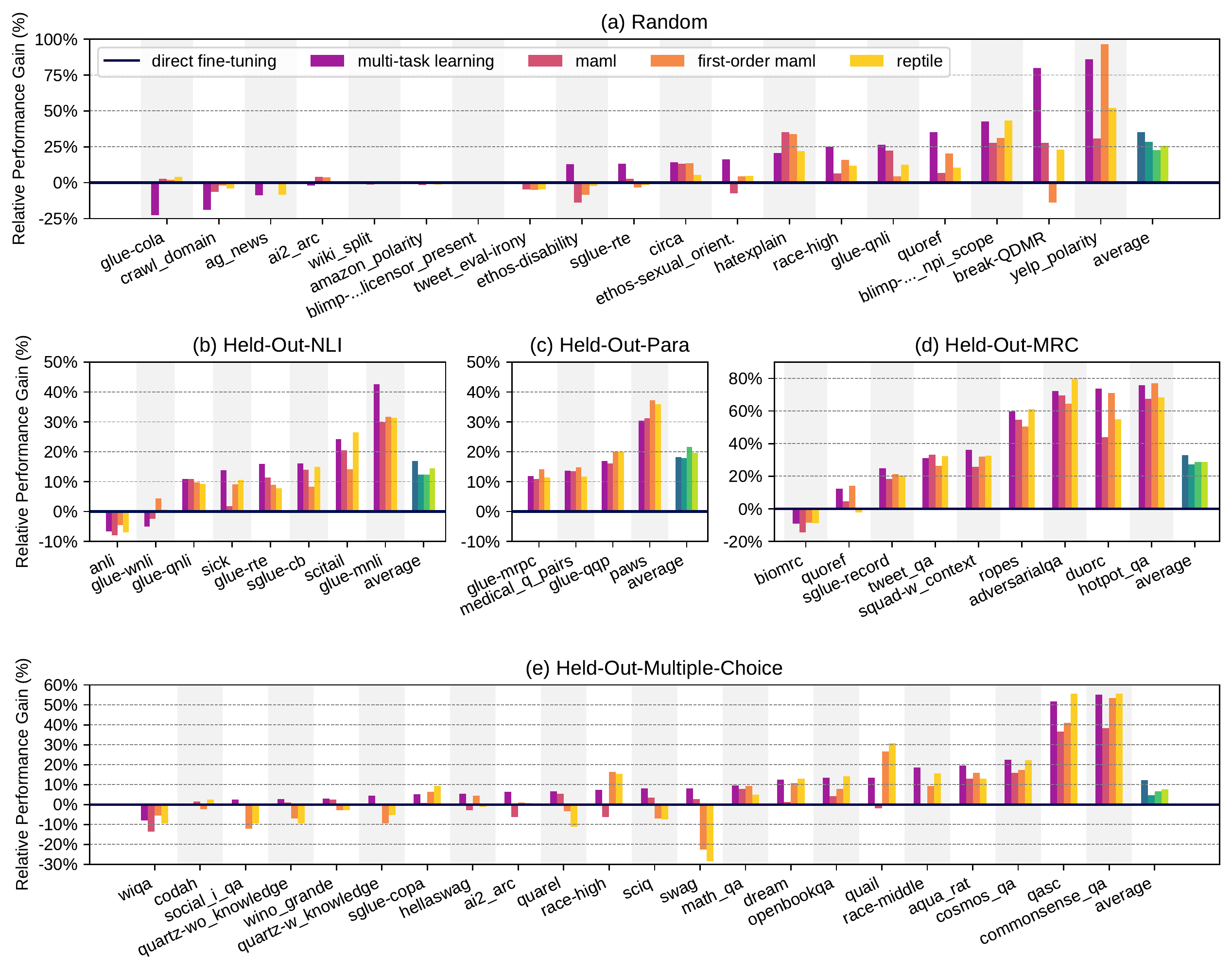}
    \caption{Experimental results for the \challengename{} challenge with different task partitions. The details of each partition is shown in Table~\ref{tab:splits}. Relative performance gain is computed based on the results of \textit{direct fine-tuning}. Best viewed in color. Green color is used to highlight the Average Relative Gain (ARG) for each method.
    }
    \label{fig:all}
\end{figure*}

\begin{figure}
    \centering
    \includegraphics[width=0.5\textwidth]{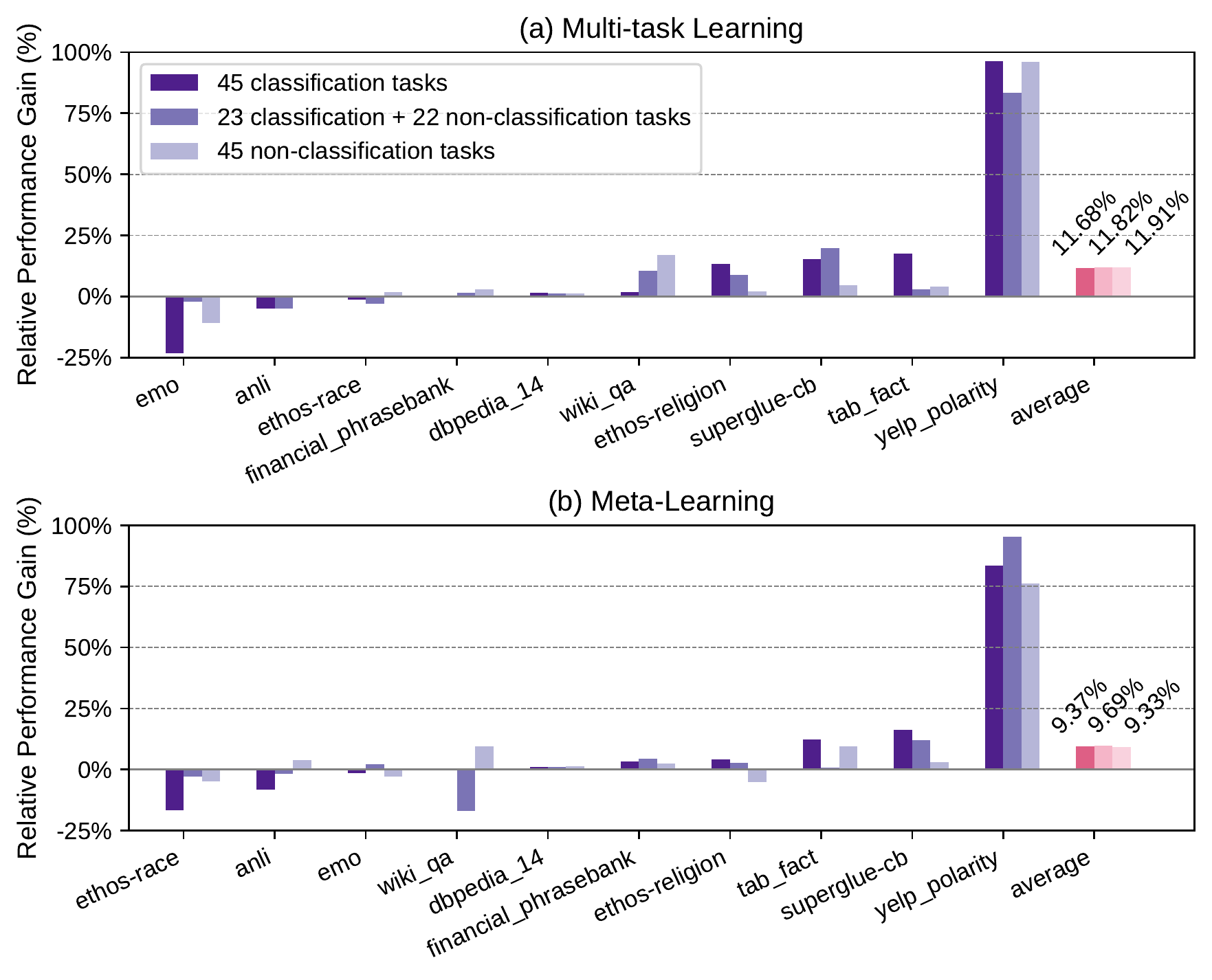}
    \caption{Comparison for the controlled experiment on Partition 2.1-2.3. $\mathcal{T}_{test}$ is a fixed set of 10 classification tasks, while $\mathcal{T}_{train}$ varies.}\vspace{-0cm}
    \label{fig:classification-merge-test}
\end{figure}

In this section we look to interpret the results and answer our research questions.
We summarize the ARG scores in Table \ref{tab:splits} and plot the performance of each test task (for each partition) in Fig.~\ref{fig:all}-\ref{fig:classification-merge-test}.

\finding{Q1. 
Can we teach pre-trained LMs to generalize across tasks with existing methods?
}


\vspace{-0.2cm}
\paragraph{Overall Performance.}
From Table \ref{tab:splits}, we observe that, on average, the tested upstream learning methods indeed improve cross-task generalization: 
their ARG scores are positive, meaning that they are better than \textit{direct fine-tuning} (ARG=0\%). 
Further, by aggregating results from all upstream learning methods and task partitions, 
we find that the performance on 51.47\% test tasks are significantly improved ($>5\%$ relative improvement compared to direct fine-tuning); 35.93\% tasks are relatively unaffected (between $\pm 5\%$); and 12.60\% tasks suffer from worse performance ($<-5\%$).



\paragraph{Correlated Performance Gains.} The performance gain obtained with different upstream learning methods are correlated with each other~--~\textit{i.e.}, tasks that benefit from multi-task learning is likely to also benefit from meta-learning. 
For the \textit{Random partition}, the Spearman Correlation between the relative improvement brought by MTL and MAML is $0.66$, 
with $p$ value equals to $0.0015$. 
This suggests that different upstream learning methods, while taking different optimization objectives, capture similar inductive bias from $\mathcal{T}_{train}$.


\paragraph{MTL is a strong baseline.} 
Surprisingly, the most straight-forward multi-task learning method is hard to beat.
This could be counter-intuitive, as meta-learning methods are specifically designed for rapid generalization to unseen tasks, 
sharing the same goal with our \challengename{} challenge. 
We think there are three possible reasons: 
(1) Due to memory constraints, we limit the number of inner-loop updates to be one, which may be insufficient. Also, meta-learning methods are highly sensitive to hyper-parameters and even random seeds \cite{antoniou2018how}, which we do not tune exhaustively for practical reasons.
(2) Text-to-text transformers have much more complex architectures, while most meta-learning methods are typically applied to small feed-forward/convolutional networks. 
(3) The \challengename{} challenge has a highly diverse set upstream tasks, which may introduce under-explored difficulties.
That being said, we believe it is important to identify the true cause, and to develop improved meta-learning methods for the \challengename{} challenge as future work.




\paragraph{Forgetting Pre-Trained Knowledge.}
A few test tasks have negative performance gain after upstream learning, including \texttt{Glue-COLA} (measuring linguistic acceptability) and \texttt{Domain Crawl} (separating domain names into tokens) in the Random Partition setting. 
For \texttt{Glue-COLA}, similar observations are reported by \citet{pruksachatkun-etal-2020-intermediate} in an intermediate-task transfer learning setting, where the authors conjecture \textit{catastrophic forgetting} of the masked language modeling (MLM) tasks may be the cause. 
BART uses denoising pre-training objective, a variant of MLM. 
Intuitively, \texttt{Domain Crawl} is also one of the most similar tasks to denoising in all test tasks, which further supports this hypothesis. 
We thus conjecture that for test tasks that resemble pre-training objectives, upstream learning could hurt performance due to the \textit{catastrophic forgetting} phenomena.

Understanding negative transfer \cite{Wu2020Understanding} and selecting source tasks to avoid negative transfer \cite{vu-etal-2020-exploring} are also growing research topics. In this work we refrain from further investigation; however we believe combating negative transfer and thus improving \challengename{} performance is a promising future direction.

\finding{Q2. Well-rounded or specialized? Which is a better strategy of upstream learning?}

``Learning to be well-rounded vs. learning to be specialized'' is a common dilemma that human learners struggles with.
For the \challengename{} challenge, the former refers to learning from a set of diverse tasks in upstream learning;
the latter refers to learning from a set of tasks closer to target few-shot tasks. 
To study this research question, we want to find out which option works better in upstream learning.
Put differently, we aim to \textbf{analyze the influence of upstream task selection} for a fixed set of the downstream tasks.

\paragraph{Setup.}
We first conduct controlled experiments with \textit{Partition 2.1-2.3}, 
where $\mathcal{T}_{test}$ is a fixed set of classification tasks, and $\mathcal{T}_{train}$ varies. 
In Partition 2.1, all tasks in $\mathcal{T}_{train}$ are \textit{classification} tasks (\textit{i.e.}, ``specialized and targeted''); 
in Partition 2.2, half of the tasks are \textit{classification} tasks (\textit{i.e.}, ``well-rounded''); 
in Partition 2.3, all tasks are \textit{non-classification} tasks (\textit{i.e.}, ``specialized in an opposite direction'', for a controlled experiment).

\paragraph{Analysis and Discussion.} 
It is surprising at first that non-classification tasks and classification tasks are equivalently helpful in terms of ARG scores (see Fig.~\ref{fig:classification-merge-test}).
On a second thought, this observation is encouraging as it demonstrates that acquiring cross-task generalization is feasible and promising, even when $\mathcal{T}_{train}$ and $\mathcal{T}_{test}$ are drastically different.
It also suggests that our categorization of tasks (\S\ref{ssec:partitions}) may not align with how models learn transferable skills: selecting $\mathcal{T}_{train}$ tasks that have the same format and goal as the test task may not lead to optimal transfer.

In retrospect, we acknowledge that our design of ontology and partitions based on task format and goal is flawed. This is merely one aspect of ``task similarity''. However, understanding the complex relationship between tasks is another challenging and under-explored problem. We consider our ontology as a starting point, rather than a fixed final one. We use the current ontology to guide our experiment and analysis, and we hope future analysis could help build a more informative ontology.


\begin{table}[t]
\centering
\scalebox{0.75}{
\begin{tabular}{cccc}
\toprule
\textbf{Test Task} & \textbf{Partition} & $\Delta_{multi}$ & $\Delta_{meta}$  \\
\midrule
\multirow{2}{*}{Glue-QNLI}  & Random & $15.89\%$ & $11.55\%$      \\
& Held-Out-NLI & $10.88\%$ & $10.94\%$ \\
\midrule
\multirow{2}{*}{AI2\_ARC}  & Random & $1.30\%$ & $4.22\%$            \\
& Held-Out-MCQA & $6.49\%$ & $-6.22\%$\\
\midrule
\multirow{2}{*}{Race-High}  & Random & $26.71\%$ & $6.59\%$      \\
& Held-Out-MCQA & $7.27\%$ & $-6.28\%$ \\
\midrule
\multirow{2}{*}{QuoRef}  & Random & $25.47\%$  & $3.99\%$     \\
& Held-Out-MRC & $12.25\%$ & $4.64\%$ \\
\bottomrule
\end{tabular}
}
\caption{Performance comparison of test task performance when different $\mathcal{T}_{train}$ sets are used in upstream learning. See text in Q2 for in-depth analysis.}\label{tab:q2}
\end{table}

\paragraph{Case Studies.}
We further look at cases where a test task appear in $\mathcal{T}_{test}$ of multiple partitions.
For example, \texttt{AI2\_ARC} and \texttt{Race-High} are in the $\mathcal{T}_{test}$ of both Random partition and Held-out-MCQA partition. 
We present the results in Table~\ref{tab:q2}. 
In general, the performance of these tasks varies when different $\mathcal{T}_{train}$ sets are used. 
However, we have not found consistent patterns of what type of $\mathcal{T}_{train}$ lead to better performance for a specific test task.






\begin{figure}
    \centering
    \includegraphics[width=0.4\textwidth]{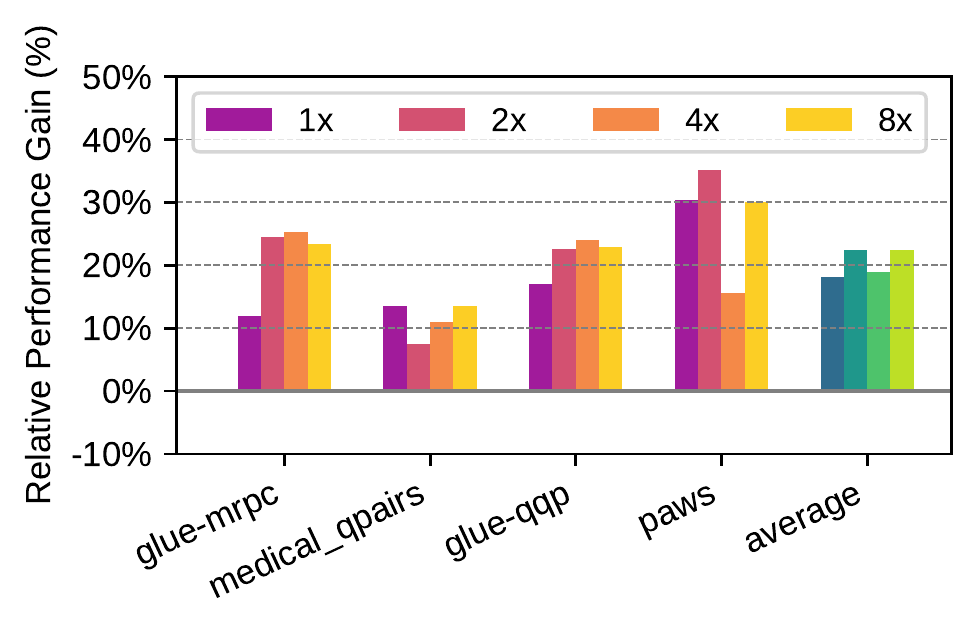}
    \caption{Controlling upstream learning data size in with Held-out-Para Partition. Enlarging the size of data during upstream learning \textit{does not} necessitate better cross-task generalization ability.}
    \label{fig:248x}
\end{figure}

\finding{Q3. Does it help if we have more labelled data for upstream tasks?} 

As described in \S\ref{ssec:sampling},
we limit our upstream tasks to be also few-shot: 
classification tasks have 16 examples per class, and non-classification tasks have 32 examples.
This decision is empirically determined following prior works \cite{schick2020exploiting,schick2020small,Gao2020MakingPL} and makes our extensive analysis practical and efficient. 
It is possible that using more data for each upstream task can significantly improve cross-task generalization.
To investigate this, 
we conduct a set of controlled experiments where the number of examples in upstream tasks are changed to $[2, 4, 8]$ times of the original size. 
We use the Held-out-Para Partition and multi-task learning for the experiments, and present the result in Fig.~\ref{fig:248x}. Surprisingly, we find that the effect from using more upstream data is inconsistent on different target tasks.
The overall ARG for all sizes are close: even 8x larger upstream data leads to only 4\% improvement in ARG.
We conclude that enlarging the size of data during upstream learning \textit{does not} necessitate better cross-task generalization ability.
This also justifies our decision to keep upstream tasks few-shot.

\finding{Q4-Q6. Additional Analysis}
Due to space limit, we summarize our other findings below and defer the details to Appendix~\ref{app:additional_exp}.

\paragraph{Few-Shot $\rightarrow$ More-Shot (Q4).}
In practice, users may continue to collect data over time.
We wonder if cross-task generalization ability is still helpful for medium/high-resource target tasks. We find that the performance gain from upstream learning is still evident when 1024 shots are available. The performance gap diminishes with millions of training examples.

\paragraph{Using Different Base Models (Q5).}
We extend our analysis on BART-base (139M) to larger pre-trained text-to-text Transformers: BART-Large (406M) and T5-v1.1-Base (248M). 
Generally, the performance grows with models sizes with only few exceptions, which suggests that upstream learning methods we use are model-agnostic, and can be applied to larger models to further improve few-shot performance. 

\paragraph{Integration with PET Training (Q6).}
Pattern-exploiting training (PET) \cite{schick2020exploiting, schick2020small} was originally proposed for classification tasks and \textit{encoder} language models. We test a few variants of PET training with BART-Base and try applying PET training after upstream learning. In general we observe deteriorated performance compared to direct fine-tuning. We hypothesize that PET methods are not directly applicable to \textit{encoder-decoder} language models used in our study. 





\section{Conclusion and Future Work}

In this paper, we study the problem of building better few-shot learners via acquiring cross-task generalization ability from diverse NLP tasks. Towards our goal, we introduce the \challengename{} Challenge, an task setup that standardizes the training pipeline, data access and evaluation protocol. We also present the NLP Few-shot Gym, a repository of 160 diverse few-shot NLP tasks, to support \challengename{} learning in different scenarios. We empirically demonstrated that cross-task generalization can be acquired via multi-task learning and meta-learning; confirmed that the selection of seen tasks would influence the few-shot performance on unseen tasks. 

We have highlighted several unexpected or undesired observations in our analysis, for which we invite future work in understanding and combating related issues. In addition, we envision the \challengename{} Challenge and the \benchmarkname{} to serve as the testbed for many interesting ``meta-problems'', such as (1) learning to generate prompt for diverse task formats and further improve learning efficiency \cite{shin-etal-2020-autoprompt,Gao2020MakingPL}; (2) learning to select appropriate source tasks to learn from during upstream learning \cite{taskonomy2018, Standley2020WhichTS}, potentially with task2vec methods \cite{Achille2019Task2VecTE, vu-etal-2020-exploring}; (3) applying task augmentation strategies to prevent over-fitting \cite{Murty2021DReCaAG}; (4) learning to accumulate knowledge and avoid catastrophic forgetting in an continual learning setup \cite{Jin2021LifelongLO}; (5) decomposing complex tasks into atomic tasks and exploring cross-task generalization through the lens of compositionality \cite{andreas-etal-2016-learning,khot-etal-2021-text}.



\section*{Acknowledgments}
We thank authors and crowd-workers of all datasets used in our study. We thank huggingface datasets team for making datasets more accessible. We thank anonymous reviewers and members of USC INK Lab for their valuable feedback. This work is supported in part by the Office of the Director of National Intelligence (ODNI), Intelligence Advanced Research Projects Activity (IARPA), via Contract No. 2019-19051600007; the DARPA MCS program under Contract No. N660011924033; the Defense Advanced Research Projects Agency with award W911NF-19-20271; NSF IIS 2048211.


\bibliography{anthology,emnlp2020,nlp_fewshot_gym_old}
\bibliographystyle{acl_natbib}

\clearpage
\appendix
\clearpage
\onecolumn
\vspace{0.5cm}


\section{Selected Tasks in NLP Few-shot Gym}
\label{sec:all_tasks_table}
\scriptsize

\begin{longtable}{lllll}
\caption{Tasks in NLP Few-shot Gym.}\label{tab:ontology}\\
\toprule
\textbf{Task Name} & \textbf{Ontology} & \textbf{Reference} \\
\midrule
\endfirsthead
\toprule
\textbf{Task Name} & \textbf{Ontology} & \textbf{Reference} \\
\midrule
\endhead
\bottomrule \multicolumn{4}{r}{Continued on next page} \\
\endfoot
\endlastfoot
acronym\_identification &	other &	\citealt{pouran-ben-veyseh-etal-2020-acronym}	\\
ade\_corpus\_v2-classification &	cls/other &	\citealt{GURULINGAPPA2012885}	\\
ade\_corpus\_v2-dosage &	other/slot filling & \citealt{GURULINGAPPA2012885}	\\
ade\_corpus\_v2-effect &	other/slot filling & \citealt{GURULINGAPPA2012885}	\\
adversarialqa &	qa/machine reading comprehension &	\citealt{bartolo-etal-2020-beat}	\\
aeslc &	cg/summarization & \citealt{zhang-tetreault-2019-email}	\\
ag\_news &	cls/topic &	\href{http://groups.di.unipi.it/~gulli/AG_corpus_of_news_articles.html}{Gulli (link)}	\\
ai2\_arc &	qa/multiple-choice qa &	\citealt{Clark2018ThinkYH}	\\
amazon\_polarity &	cls/sentiment analysis & \citealt{McAuley2013HiddenFA}	\\
anli &	cls/nli & \citealt{nie-etal-2020-adversarial}	\\
app\_reviews &	other/regression &	Missing\\ 
aqua\_rat &	qa/multiple-choice qa &	\citealt{ling-etal-2017-program}	\\
art (abductive nli) &	other &	\citealt{bhagavatula2020abductive}	\\
aslg\_pc12 &	other &	\citealt{Othman2012EnglishASLGP}	\\
biomrc &	qa/machine reading comprehension &	\citealt{pappas-etal-2020-biomrc}	\\
blimp-anaphor\_gender\_agreement &	other/linguistic phenomenon & \citealt{warstadt2019blimp}	\\
blimp-anaphor\_number\_agreement &	other/linguistic phenomenon & \citealt{warstadt2019blimp}	\\
blimp-determiner\_noun\_agreement\_with\_adj\_irregular\_1 &	other/linguistic phenomenon & \citealt{warstadt2019blimp}	\\
blimp-ellipsis\_n\_bar\_1 &	other/linguistic phenomenon & \citealt{warstadt2019blimp}	\\
blimp-ellipsis\_n\_bar\_2 &	other/linguistic phenomenon & \citealt{warstadt2019blimp}	\\
blimp-existential\_there\_quantifiers\_1 &	other/linguistic phenomenon & \citealt{warstadt2019blimp}	\\
blimp-irregular\_past\_participle\_adjectives &	other/linguistic phenomenon & \citealt{warstadt2019blimp}	\\
blimp-sentential\_negation\_npi\_licensor\_present &	other/linguistic phenomenon & \citealt{warstadt2019blimp}	\\
blimp-sentential\_negation\_npi\_scope &	other/linguistic phenomenon & \citealt{warstadt2019blimp}	\\
blimp-wh\_questions\_object\_gap &	other/linguistic phenomenon & \citealt{warstadt2019blimp}	\\
boolq &	qa/binary &	\citealt{clark-etal-2019-boolq}	\\
break-QDMR &	other &	\citealt{wolfson-etal-2020-break}	\\
break-QDMR-high-level &	other & \citealt{wolfson-etal-2020-break}	\\
circa &	cls/other &	\citealt{louis-etal-2020-id}	\\
climate\_fever &	cls/fact checking &	\citealt{Diggelmann2020CLIMATEFEVERAD}	\\
codah &	qa/multiple-choice qa &	\citealt{chen-etal-2019-codah}	\\
common\_gen &	other &	\citealt{lin-etal-2020-commongen}	\\
commonsense\_qa &	qa/multiple-choice qa &	\citealt{talmor-etal-2019-commonsenseqa}	\\
cos\_e &	other/generate explanation & \citealt{rajani-etal-2019-explain}	\\
cosmos\_qa &	qa/multiple-choice qa &	\citealt{huang-etal-2019-cosmos}	\\
crawl\_domain &	other &	\citealt{zhang-etal-2020-semi}	\\
crows\_pairs &	other &	\citealt{nangia-etal-2020-crows}	\\
dbpedia\_14 &	cls/topic &	\citealt{Lehmann2015DBpediaA}	\\
definite\_pronoun\_resolution &	other &	\citealt{rahman-ng-2012-resolving}	\\
discovery &	cls/other &	\citealt{sileo-etal-2019-mining}	\\
dream &	qa/multiple-choice qa &	\citealt{sun-etal-2019-dream}	\\
duorc &	qa/machine reading comprehension &	\citealt{saha-etal-2018-duorc}	\\
e2e\_nlg\_cleaned &	other &	\citealt{dusek.etal2020:csl, dusek-etal-2019-semantic}	\\
eli5-askh &	qa/long-form qa & \citealt{fan-etal-2019-eli5}	\\
eli5-asks &	qa/long-form qa & \citealt{fan-etal-2019-eli5}	\\
eli5-eli5 &	qa/long-form qa & \citealt{fan-etal-2019-eli5}	\\
emo &	cls/emotion & \citealt{chatterjee-etal-2019-semeval}	\\
emotion &	cls/emotion & \citealt{saravia-etal-2018-carer}	\\
empathetic\_dialogues &	cg/dialogue & \citealt{rashkin-etal-2019-towards}	\\
ethos-directed\_vs\_generalized &	cls/hate speech detection &	\citealt{Mollas2020ETHOSAO}	\\
ethos-disability &	cls/hate speech detection &	\citealt{Mollas2020ETHOSAO}	\\
ethos-gender &	cls/hate speech detection &	\citealt{Mollas2020ETHOSAO}	\\
ethos-national\_origin &	cls/hate speech detection &	\citealt{Mollas2020ETHOSAO}	\\
ethos-race &	cls/hate speech detection &	\citealt{Mollas2020ETHOSAO}	\\
ethos-religion &	cls/hate speech detection &	\citealt{Mollas2020ETHOSAO}	\\
ethos-sexual\_orientation &	cls/hate speech detection &	\citealt{Mollas2020ETHOSAO}	\\
financial\_phrasebank &	cls/sentiment analysis &	\citealt{financial-phrasebank}	\\
freebase\_qa &	qa/closed-book qa &	\citealt{jiang-etal-2019-freebaseqa}	\\
gigaword &	cg/summarization &	\citealt{napoles-etal-2012-annotated}	\\
glue-cola &	cls/other &	\citealt{warstadt-etal-2019-neural}	\\
glue-mnli &	cls/nli & \citealt{williams-etal-2018-broad}	\\
glue-mrpc &	cls/paraphrase & \citealt{dolan-brockett-2005-automatically}\\
glue-qnli &	cls/nli &	\citealt{rajpurkar-etal-2016-squad}	\\
glue-qqp &	cls/paraphrase & \href{http://data.quora.com/First-Quora-Dataset-Release-Question-Pairs}{(link)}	\\
glue-rte &	cls/nli &	\begin{tabular}[c]{@{}l@{}}\citealt{dagan2005pascal, bar2006second}\\\citealt{giampiccolo2007third, bentivogli2009fifth}\end{tabular}	\\
glue-sst2 &	cls/sentiment analysis &	\citealt{socher-etal-2013-recursive}	\\
glue-wnli &	cls/nli & \citealt{levesque2012winograd}	\\
google\_wellformed\_query &	cls/other &	\citealt{faruqui-das-2018-identifying}	\\
hate\_speech18 &	cls/hate speech detection &	\citealt{gibert2018hate}	\\
hate\_speech\_offensive &	cls/hate speech detection &	\citealt{hateoffensive}	\\
hatexplain &	cls/hate speech detection &	\citealt{mathew2020hatexplain}	\\
health\_fact &	cls/fact checking &	\citealt{kotonya-toni-2020-explainable-automated}	\\
hellaswag &	qa/multiple-choice qa &	\citealt{zellers-etal-2019-hellaswag}	\\
hotpot\_qa &	qa/machine reading comprehension &	\citealt{yang-etal-2018-hotpotqa}	\\
imdb &	cls/sentiment analysis & \citealt{maas-etal-2011-learning}	\\
jeopardy &	qa/closed-book qa &	\href{https://www.reddit.com/r/datasets/comments/1uyd0t/200000_jeopardy_questions_in_a_json_file/}{(link)}	\\
kilt\_ay2 &	other/entity linking &	\citealt{hoffart-etal-2011-robust}	\\
kilt\_fever &	cls/fact checking &	\citealt{thorne-etal-2018-fever}	\\
kilt\_hotpotqa &	qa/closed-book qa &	\citealt{yang-etal-2018-hotpotqa}	\\
kilt\_nq &	qa/closed-book qa &	\citealt{kwiatkowski-etal-2019-natural}	\\
kilt\_trex &	qa/closed-book qa &	\citealt{elsahar-etal-2018-rex}	\\
kilt\_wow &	cg/dialogue &	\citealt{dinan2018wizard}	\\
kilt\_zsre &	qa/closed-book qa &	\citealt{levy-etal-2017-zero}	\\
lama-conceptnet &	qa/closed-book qa &	\citealt{petroni-etal-2019-language,petroni2020how}	\\
lama-google\_re &	qa/closed-book qa &	\citealt{petroni-etal-2019-language,petroni2020how}	\\
lama-squad &	qa/closed-book qa &	\citealt{petroni-etal-2019-language,petroni2020how}	\\
lama-trex &	qa/closed-book qa &	\citealt{petroni-etal-2019-language,petroni2020how}	\\
liar &	cls/fact checking &	\citealt{wang-2017-liar}	\\
limit &	other &	\citealt{manotas-etal-2020-limit}	\\
math\_qa &	qa/multiple-choice qa &	\citealt{amini-etal-2019-mathqa}	\\
mc\_taco &	qa/binary &	\citealt{zhou-etal-2019-going}	\\
medical\_questions\_pairs &	cls/paraphrase & \citealt{medical-qqp}	\\
mocha &	other/regression &	\citealt{chen-etal-2020-mocha}	\\
multi\_news &	cg/summarization & \citealt{fabbri-etal-2019-multi}	\\
numer\_sense &	qa/closed-book qa &	\citealt{lin-etal-2020-birds}	\\
onestop\_english &	cls/other &	\citealt{vajjala-lucic-2018-onestopenglish}	\\
openbookqa &	qa/multiple-choice qa &	\citealt{mihaylov-etal-2018-suit}	\\
paws &	cls/paraphrase & \citealt{zhang-etal-2019-paws}	\\
piqa &	other &	\citealt{Bisk2020}	\\
poem\_sentiment &	cls/sentiment analysis & \citealt{sheng-uthus-2020-investigating}	\\
proto\_qa &	other &	\citealt{boratko-etal-2020-protoqa}	\\
qa\_srl &	other &	\citealt{he-etal-2015-question}	\\
qasc &	qa/multiple-choice qa &	\citealt{Khot_Clark_Guerquin_Jansen_Sabharwal_2020}	\\
quail &	qa/multiple-choice qa &	\citealt{Rogers_Kovaleva_Downey_Rumshisky_2020}	\\
quarel &	qa/multiple-choice qa &	\citealt{Tafjord_Clark_Gardner_Yih_Sabharwal_2019}	\\
quartz-no\_knowledge &	qa/multiple-choice qa &	\citealt{tafjord-etal-2019-quartz}	\\
quartz-with\_knowledge &	qa/multiple-choice qa &	\citealt{tafjord-etal-2019-quartz}	\\
quoref &	qa/machine reading comprehension &	\citealt{dasigi-etal-2019-quoref}	\\
race-high &	qa/multiple-choice qa &	\citealt{lai-etal-2017-race}	\\
race-middle &	qa/multiple-choice qa &	\citealt{lai-etal-2017-race}	\\
reddit\_tifu-title &	cg/summarization &	\citealt{kim-etal-2019-abstractive}	\\
reddit\_tifu-tldr &	cg/summarization &	\citealt{kim-etal-2019-abstractive}	\\
ropes &	qa/machine reading comprehension &	\citealt{lin-etal-2019-reasoning}	\\
rotten\_tomatoes &	cls/sentiment analysis & \citealt{pang-lee-2005-seeing}	\\
samsum &	cg/summarization &	\citealt{gliwa-etal-2019-samsum}	\\
scicite &	cls/other &	\citealt{cohan-etal-2019-structural}	\\
sciq &	qa/multiple-choice qa &	\citealt{welbl-etal-2017-crowdsourcing}	\\
scitail &	cls/nli & \citealt{scitail}	\\
search\_qa &	qa/closed-book qa &	\citealt{Dunn2017SearchQAAN}	\\
sick &	cls/nli &	\citealt{marelli-etal-2014-sick}	\\
sms\_spam &	cls/other &	\citealt{sms_spam}	\\
social\_i\_qa &	qa/multiple-choice qa &	\citealt{sap-etal-2019-social}	\\
spider &	cg/other &	\citealt{yu-etal-2018-spider}	\\
squad-no\_context &	qa/closed-book qa &	\citealt{rajpurkar-etal-2016-squad}	\\
squad-with\_context &	qa/machine reading comprehension &	\citealt{rajpurkar-etal-2016-squad}	\\
superglue-cb &	cls/nli & \citealt{Marneffe_Simons_Tonhauser_2019}	\\
superglue-copa &	qa/multiple-choice qa &	\citealt{gordon-etal-2012-semeval} \\
superglue-multirc &	qa/multiple-choice qa &	\citealt{khashabi-etal-2018-looking}	\\
superglue-record &	qa/machine reading comprehension & \citealt{Zhang2018ReCoRDBT}	\\
superglue-rte &	cls/nli & \begin{tabular}[c]{@{}l@{}}\citealt{dagan2005pascal, bar2006second}\\\citealt{giampiccolo2007third, bentivogli2009fifth}\end{tabular}	\\
superglue-wic &	cls/other &	\citealt{pilehvar-camacho-collados-2019-wic}	\\
superglue-wsc &	cls/other &	\citealt{levesque2012winograd}	\\
swag &	qa/multiple-choice qa &	\citealt{zellers-etal-2018-swag}	\\
tab\_fact &	cls/fact checking &	\citealt{Chen2020TabFact}	\\
trec &	cls/other &	\citealt{li-roth-2002-learning,hovy-etal-2001-toward}	\\
trec-finegrained &	cls/other &	\citealt{li-roth-2002-learning,hovy-etal-2001-toward}	\\
tweet\_eval-emoji &	cls/emotion & \citealt{barbieri-etal-2020-tweeteval}	\\
tweet\_eval-emotion &	cls/emotion &	\citealt{barbieri-etal-2020-tweeteval}	\\
tweet\_eval-hate &	cls/emotion &	\citealt{barbieri-etal-2020-tweeteval}	\\
tweet\_eval-irony &	cls/emotion &	\citealt{barbieri-etal-2020-tweeteval}	\\
tweet\_eval-offensive &	cls/emotion &	\citealt{barbieri-etal-2020-tweeteval}	\\
tweet\_eval-sentiment &	cls/emotion &	\citealt{barbieri-etal-2020-tweeteval}	\\
tweet\_eval-stance\_abortion &	cls/emotion &	\citealt{barbieri-etal-2020-tweeteval}	\\
tweet\_eval-stance\_atheism &	cls/emotion &	\citealt{barbieri-etal-2020-tweeteval}	\\
tweet\_eval-stance\_climate &	cls/emotion &	\citealt{barbieri-etal-2020-tweeteval}	\\
tweet\_eval-stance\_feminist &	cls/emotion &	\citealt{barbieri-etal-2020-tweeteval}	\\
tweet\_eval-stance\_hillary &	cls/emotion &	\citealt{barbieri-etal-2020-tweeteval}	\\
tweet\_qa &	qa/machine reading comprehension &	\citealt{xiong-etal-2019-tweetqa}	\\
web\_questions &	qa/closed-book qa &	\citealt{berant-etal-2013-semantic}	\\
wiki\_auto &	cls/other &	\citealt{jiang-etal-2020-neural}	\\
wiki\_bio &	cg/other &	\citealt{lebret-etal-2016-neural}	\\
wiki\_qa &	cls/other &	\citealt{yang-etal-2015-wikiqa}	\\
wiki\_split &	cg/other & \citealt{botha-etal-2018-learning}	\\
wikisql &	cg/other &	\citealt{zhongSeq2SQL2017}	\\
wino\_grande &	qa/multiple-choice qa &	\citealt{Sakaguchi_Le_Bras_Bhagavatula_Choi_2020}	\\
wiqa &	qa/multiple-choice qa &	\citealt{tandon-etal-2019-wiqa}	\\
xsum &	cg/summarization &	\citealt{narayan-etal-2018-dont}	\\
yahoo\_answers\_topics &	cls/topic &	\href{https://webscope.sandbox.yahoo.com/catalog.php?datatype=l}{(link)}	\\
yelp\_polarity &	cls/sentiment analysis & \citealt{zhang2015character}; \href{https://www.yelp.com/dataset}{(link)}	\\
yelp\_review\_full &	other/regression &	\citealt{zhang2015character}; \href{https://www.yelp.com/dataset}{(link)}	\\
\bottomrule
\end{longtable}
\normalsize



\section{Details about Task Partition}

\label{app:partitions}
\subsection{Partition 1. Random}

\begin{lstlisting}[language=json,firstnumber=1]
{
    "train": ['glue-mrpc', 'math_qa', 'quarel', 'e2e_nlg_cleaned', 'tweet_eval-stance_atheism', 'lama-squad', 'tab_fact', 'aqua_rat', 'tweet_eval-emoji', 'glue-wnli', 'codah', 'tweet_eval-offensive', 'wiki_qa', 'blimp-ellipsis_n_bar_1', 'openbookqa', 'sms_spam', 'acronym_identification', 'blimp-determiner_noun_agreement_with_adj_irregular_1', 'ethos-national_origin', 'spider', 'definite_pronoun_resolution', 'hellaswag', 'superglue-wsc', 'numer_sense', 'ade_corpus_v2-dosage', 'blimp-ellipsis_n_bar_2', 'kilt_ay2', 'squad-no_context', 'google_wellformed_query', 'xsum', 'wiqa', 'tweet_eval-stance_abortion', 'reddit_tifu-tldr', 'ade_corpus_v2-effect', 'qa_srl', 'ethos-religion', 'commonsense_qa', 'jeopardy', 'biomrc', 'superglue-multirc', 'ethos-race', 'eli5-askh', 'glue-qqp', 'paws', 'ethos-directed_vs_generalized', 'glue-sst2', 'mocha', 'tweet_eval-hate', 'glue-rte', 'blimp-anaphor_number_agreement', 'lama-conceptnet', 'hate_speech_offensive', 'superglue-wic', 'boolq', 'kilt_hotpotqa', 'quartz-no_knowledge', 'aslg_pc12', 'sick', 'tweet_eval-stance_climate', 'tweet_eval-sentiment', 'crows_pairs', 'glue-mnli', 'medical_questions_pairs', 'break-QDMR-high-level', 'qasc', 'imdb', 'ethos-gender', 'trec-finegrained', 'adversarialqa', 'onestop_english', 'web_questions', 'duorc', 'yelp_review_full', 'swag', 'proto_qa', 'scitail', 'tweet_eval-stance_feminist', 'limit', 'common_gen', 'scicite', 'blimp-irregular_past_participle_adjectives', 'social_i_qa', 'anli', 'kilt_zsre', 'cosmos_qa', 'superglue-record', 'squad-with_context', 'emotion', 'blimp-existential_there_quantifiers_1', 'race-middle', 'kilt_wow', 'sciq', 'wino_grande', 'rotten_tomatoes', 'superglue-cb', 'poem_sentiment', 'ropes', 'reddit_tifu-title', 'piqa', 'climate_fever', 'lama-google_re', 'search_qa', 'wiki_auto', 'mc_taco', 'blimp-wh_questions_object_gap', 'hotpot_qa', 'emo', 'kilt_nq', 'kilt_trex', 'quartz-with_knowledge', 'dbpedia_14', 'yahoo_answers_topics', 'app_reviews', 'superglue-copa', 'blimp-anaphor_gender_agreement', 'hate_speech18', 'gigaword', 'multi_news', 'aeslc', 'quail'],
    "dev": ['cos_e', 'kilt_fever', 'eli5-asks', 'trec', 'eli5-eli5', 'art', 'empathetic_dialogues', 'tweet_qa', 'wikisql', 'lama-trex', 'tweet_eval-stance_hillary', 'discovery', 'tweet_eval-emotion', 'liar', 'wiki_bio', 'dream', 'ade_corpus_v2-classification', 'health_fact', 'samsum', 'financial_phrasebank'],
    "test": ['quoref', 'wiki_split', 'ethos-disability', 'yelp_polarity', 'superglue-rte', 'glue-cola', 'ethos-sexual_orientation', 'blimp-sentential_negation_npi_scope', 'ai2_arc', 'amazon_polarity', 'race-high', 'blimp-sentential_negation_npi_licensor_present', 'tweet_eval-irony', 'break-QDMR', 'crawl_domain', 'freebase_qa', 'glue-qnli', 'hatexplain', 'ag_news', 'circa'],
}

\end{lstlisting}

\subsection{Partition 2.1. 45cls}

\begin{lstlisting}[language=json,firstnumber=1]
{
    "train": ["superglue-rte", "tweet_eval-sentiment", "discovery", "glue-rte", "superglue-wsc", "scicite", "glue-mrpc", "tweet_eval-stance_hillary", "tweet_eval-offensive", "emotion", "hatexplain", "glue-cola", "sick", "paws", "ethos-sexual_orientation", "glue-qqp", "tweet_eval-emotion", "sms_spam", "health_fact", "glue-mnli", "imdb", "ethos-disability", "glue-wnli", "scitail", "trec-finegrained", "yahoo_answers_topics", "liar", "glue-sst2", "tweet_eval-stance_abortion", "circa", "tweet_eval-stance_climate", "glue-qnli", "tweet_eval-emoji", "ethos-directed_vs_generalized", "ade_corpus_v2-classification", "wiki_auto", "hate_speech_offensive", "superglue-wic", "google_wellformed_query", "tweet_eval-irony", "ethos-gender", "onestop_english", "trec", "rotten_tomatoes", "kilt_fever"],
    "dev": ["tweet_eval-stance_feminist", "ethos-national_origin", "tweet_eval-hate", "ag_news", "amazon_polarity", "hate_speech18", "poem_sentiment", "climate_fever", "medical_questions_pairs", "tweet_eval-stance_atheism"],
    "test": ["superglue-cb", "dbpedia_14", "wiki_qa", "emo", "yelp_polarity", "ethos-religion", "financial_phrasebank", "tab_fact", "anli", "ethos-race"],
}
\end{lstlisting}

\subsection{Partition 2.2. 23cls+22non-cls}

\begin{lstlisting}[language=json,firstnumber=1]
{
    "train": ["ade_corpus_v2-dosage", "biomrc", "blimp-ellipsis_n_bar_2", "blimp-sentential_negation_npi_scope", "commonsense_qa", "crows_pairs", "duorc", "hellaswag", "kilt_zsre", "lama-google_re", "lama-squad", "math_qa", "numer_sense", "openbookqa", "piqa", "proto_qa", "quartz-no_knowledge", "race-high", "reddit_tifu-tldr", "ropes", "sciq", "wiki_bio", "discovery", "emotion", "ethos-disability", "ethos-sexual_orientation", "glue-cola", "glue-mnli", "glue-mrpc", "glue-qqp", "glue-rte", "glue-wnli", "hatexplain", "health_fact", "imdb", "paws", "scicite", "sick", "sms_spam", "superglue-rte", "superglue-wsc", "tweet_eval-emotion", "tweet_eval-offensive", "tweet_eval-sentiment", "tweet_eval-stance_hillary"],
    "dev": ["tweet_eval-stance_feminist", "ethos-national_origin", "tweet_eval-hate", "ag_news", "amazon_polarity", "hate_speech18", "poem_sentiment", "climate_fever", "medical_questions_pairs", "tweet_eval-stance_atheism"],
    "test": ["superglue-cb", "dbpedia_14", "wiki_qa", "emo", "yelp_polarity", "ethos-religion", "financial_phrasebank", "tab_fact", "anli", "ethos-race"]
}
\end{lstlisting}

\subsection{Partition 2.3. 45non-cls}

\begin{lstlisting}[language=json,firstnumber=1]
{
    "train": ["ade_corpus_v2-dosage", "art", "biomrc", "blimp-anaphor_number_agreement", "blimp-ellipsis_n_bar_2", "blimp-sentential_negation_npi_licensor_present", "blimp-sentential_negation_npi_scope", "break-QDMR-high-level", "commonsense_qa", "crows_pairs", "dream", "duorc", "eli5-asks", "eli5-eli5", "freebase_qa", "gigaword", "hellaswag", "hotpot_qa", "kilt_ay2", "kilt_hotpotqa", "kilt_trex", "kilt_zsre", "lama-conceptnet", "lama-google_re", "lama-squad", "math_qa", "numer_sense", "openbookqa", "piqa", "proto_qa", "qa_srl", "quarel", "quartz-no_knowledge", "race-high", "reddit_tifu-title", "reddit_tifu-tldr", "ropes", "sciq", "social_i_qa", "spider", "superglue-multirc", "wiki_bio", "wikisql", "xsum", "yelp_review_full"],
    "dev": ["tweet_eval-stance_feminist", "ethos-national_origin", "tweet_eval-hate", "ag_news", "amazon_polarity", "hate_speech18", "poem_sentiment", "climate_fever", "medical_questions_pairs", "tweet_eval-stance_atheism"],
    "test": ["superglue-cb", "dbpedia_14", "wiki_qa", "emo", "yelp_polarity", "ethos-religion", "financial_phrasebank", "tab_fact", "anli", "ethos-race"]
}
\end{lstlisting}

\subsection{Partition 3.1. Held-out-NLI}

\begin{lstlisting}[language=json,firstnumber=1]
{
    "train": ["ade_corpus_v2-classification", "ag_news", "amazon_polarity", "circa", "climate_fever", "dbpedia_14", "discovery", "emo", "emotion", "ethos-directed_vs_generalized", "ethos-disability", "ethos-gender", "ethos-national_origin", "ethos-race", "ethos-religion", "ethos-sexual_orientation", "financial_phrasebank", "glue-cola", "glue-mrpc", "glue-qqp", "glue-sst2", "google_wellformed_query", "hate_speech18", "hate_speech_offensive", "hatexplain", "health_fact", "imdb", "kilt_fever", "liar", "medical_questions_pairs", "onestop_english", "paws", "poem_sentiment", "rotten_tomatoes", "scicite", "sick", "sms_spam", "superglue-wic", "superglue-wsc", "tab_fact", "trec", "trec-finegrained", "tweet_eval-emoji", "tweet_eval-emotion", "tweet_eval-hate", "tweet_eval-irony", "tweet_eval-offensive", "tweet_eval-sentiment", "tweet_eval-stance_abortion", "tweet_eval-stance_atheism", "tweet_eval-stance_climate", "tweet_eval-stance_feminist", "tweet_eval-stance_hillary", "wiki_auto", "wiki_qa", "yahoo_answers_topics", "yelp_polarity"
    ],
    "dev": [],
    "test": ["anli", "glue-mnli", "glue-qnli", "glue-rte", "glue-wnli", "scitail", "sick", "superglue-cb"]
}
\end{lstlisting}

\subsection{Partition 3.2. Held-out-Para}

\begin{lstlisting}[language=json,firstnumber=1]
{
    "train": ["ade_corpus_v2-classification", "ag_news", "amazon_polarity", "anli", "circa", "climate_fever", "dbpedia_14", "discovery", "emo", "emotion", "ethos-directed_vs_generalized", "ethos-disability", "ethos-gender", "ethos-national_origin", "ethos-race", "ethos-religion", "ethos-sexual_orientation", "financial_phrasebank", "glue-cola", "glue-mnli", "glue-qnli", "glue-rte", "glue-sst2", "glue-wnli", "google_wellformed_query", "hate_speech18", "hate_speech_offensive", "hatexplain", "health_fact", "imdb", "kilt_fever", "liar", "onestop_english", "poem_sentiment", "rotten_tomatoes", "scicite", "scitail", "sick", "sms_spam", "superglue-cb", "superglue-rte", "superglue-wic", "superglue-wsc", "tab_fact", "trec", "trec-finegrained", "tweet_eval-emoji", "tweet_eval-emotion", "tweet_eval-hate", "tweet_eval-irony", "tweet_eval-offensive", "tweet_eval-sentiment", "tweet_eval-stance_abortion", "tweet_eval-stance_atheism", "tweet_eval-stance_climate", "tweet_eval-stance_feminist", "tweet_eval-stance_hillary", "wiki_auto", "wiki_qa", "yahoo_answers_topics", "yelp_polarity"],
    "dev": [],
    "test": ["glue-mrpc", "glue-qqp", "medical_questions_pairs", "paws"]
}

\end{lstlisting}

\subsection{Partition 4.1. Held-out-MRC}

\begin{lstlisting}[language=json,firstnumber=1]
{
    "train": ["ai2_arc", "aqua_rat", "boolq", "codah", "commonsense_qa", "cosmos_qa", "dream", "eli5-askh", "eli5-asks", "eli5-eli5", "freebase_qa", "hellaswag", "jeopardy", "kilt_hotpotqa", "kilt_nq", "kilt_trex", "kilt_zsre", "lama-conceptnet", "lama-google_re", "lama-squad", "lama-trex", "math_qa", "mc_taco", "numer_sense", "openbookqa", "qasc", "quail", "quarel", "quartz-no_knowledge", "quartz-with_knowledge", "race-high", "race-middle", "sciq", "search_qa", "social_i_qa", "squad-no_context", "superglue-copa", "superglue-multirc", "swag", "web_questions", "wino_grande", "wiqa"
    ],
    "dev": [],
    "test": ["adversarialqa", "biomrc", "duorc", "hotpot_qa", "quoref", "ropes", "squad-with_context", "superglue-record", "tweet_qa"],
}
\end{lstlisting}

\subsection{Partition 4.2. Held-out-MCQA}

\begin{lstlisting}[language=json,firstnumber=1]
{
    "train": ["adversarialqa", "biomrc", "boolq", "duorc", "eli5-askh", "eli5-asks", "eli5-eli5", "freebase_qa", "hotpot_qa", "jeopardy", "kilt_hotpotqa", "kilt_nq", "kilt_trex", "kilt_zsre", "lama-conceptnet", "lama-google_re", "lama-squad", "lama-trex", "mc_taco", "numer_sense", "quoref", "ropes", "search_qa", "squad-no_context", "squad-with_context", "superglue-multirc", "superglue-record", "tweet_qa", "web_questions"
    ],
    "dev": [],
    "test": ["ai2_arc", "aqua_rat", "codah", "commonsense_qa", "cosmos_qa", "dream", "hellaswag", "math_qa", "openbookqa", "qasc", "quail", "quarel", "quartz-no_knowledge", "quartz-with_knowledge", "race-high", "race-middle", "sciq", "social_i_qa", "superglue-copa", "swag", "wino_grande", "wiqa"]
}
\end{lstlisting}

\subsection{Partition 5. Held-out-GLUE}
To examine whether combining our methods with template-based training \cite{schick2020exploiting,schick2020small,Gao2020MakingPL} results in even better few-shot performance, we add another partition that uses all non-GLUE classification tasks as $\mathcal{T}_{train}$, and all GLUE tasks as $\mathcal{T}_{test}$.
\begin{lstlisting}[language=json,firstnumber=1]
{
    "train": ["ade_corpus_v2-classification", "ag_news", "amazon_polarity", "anli", "circa", "climate_fever", "dbpedia_14", "discovery", "emo", "emotion", "ethos-directed_vs_generalized", "ethos-disability", "ethos-gender", "ethos-national_origin", "ethos-race", "ethos-religion", "ethos-sexual_orientation", "financial_phrasebank", "google_wellformed_query", "hate_speech18", "hate_speech_offensive", "hatexplain", "health_fact", "imdb", "kilt_fever", "liar", "medical_questions_pairs", "onestop_english", "paws", "poem_sentiment", "rotten_tomatoes", "scicite", "scitail", "sick", "sms_spam", "superglue-cb", "superglue-wic", "superglue-wsc", "tab_fact", "trec", "trec-finegrained", "tweet_eval-emoji", "tweet_eval-emotion", "tweet_eval-hate", "tweet_eval-irony", "tweet_eval-offensive", "tweet_eval-sentiment", "tweet_eval-stance_abortion", "tweet_eval-stance_atheism", "tweet_eval-stance_climate", "tweet_eval-stance_feminist", "tweet_eval-stance_hillary", "wiki_auto", "wiki_qa", "yahoo_answers_topics", "yelp_polarity"],
    "dev": [],
    "test": ["glue-cola", "glue-mnli", "glue-mrpc", "glue-qnli", "glue-qqp", "glue-rte", "glue-sst2", "glue-wnli"]
}
\end{lstlisting}

Continued on next page.

\twocolumn
\section{Additional Results and Analysis}
\label{app:additional_exp}

\finding{Q4. Does the improved cross-task generalization ability go beyond few-shot settings?} 

\begin{figure*}
    \centering
    \includegraphics[width=1\textwidth]{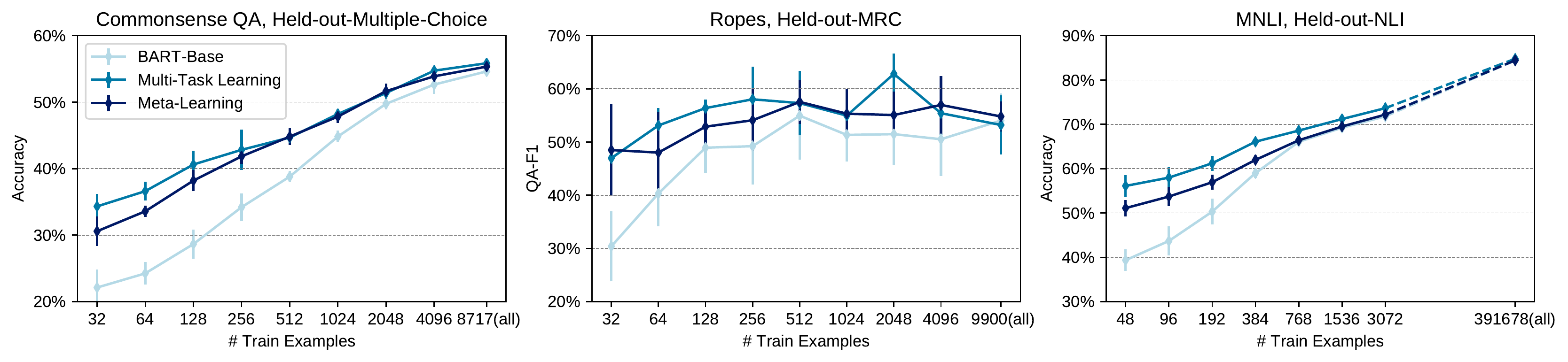}
    \caption{Performance comparisons in medium and high-resource scenarios. Benefits brought by upstream learning lasts in medium-resource scenarios.}
    \label{fig:more-shots}
\end{figure*}

In real-world applications, 
annotated data usually grow for a few-shot task over time.
Is upstream learning still helpful when a target task has more shots? 
To study this question, we study \texttt{CommonsenseQA} 
(in \textit{Held-out-Multiple-Choice Partition}), \texttt{ROPES} (in \textit{Held-out-MRC Partition}), and \texttt{MNLI} (in \textit{Held-out-NLI Partition}) as target tasks in medium and high-resource scenarios.
We take their corresponding checkpoints after upstream learning and conduct experiments 
in medium and high-resource scenarios. 
That is, we randomly sample $\{32, 64, \dots, 4096\}$ examples from the three datasets, and use them as $\mathcal{D}_{train}$. 
Then, we sample a $\mathcal{D}_{dev}$ with the same size as $\mathcal{D}_{train}$, or has the size of 1024 if $|\mathcal{D}_{train}|>1024$. 
We also try fine-tuning with the full dataset.\footnote{We do five random samples of 1024 examples as $\mathcal{D}_{dev}$ and use the remaining examples in the original train set as $\mathcal{D}_{train}$. We use the original dev set for testing.}
The performance of these settings is shown in Fig.~\ref{fig:more-shots}. 

From Fig.~\ref{fig:more-shots},
we see that the benefits brought by upstream learning methods extend into medium resource cases with up to 2048 training examples. 
For \texttt{CommonsenseQA}, checkpoints from upstream learning outperform direct fine-tuning significantly, even with the full dataset.
This finding encourages the use of upstream learning before task-specific fine-tuning when the target task has limited annotation.
On the other hand, for resource-rich tasks (e.g., MNLI), the improvement brought by upstream learning diminishes. 
This aligns with the findings of \citep{wang-etal-2020-pretrain} who discuss the benefits of pre-training on resource-rich tasks. 

\finding{Q5. Can we further improve few-shot performance by using different/larger pre-trained models?}
\begin{figure*}
    \centering
    \includegraphics[width=0.9\textwidth]{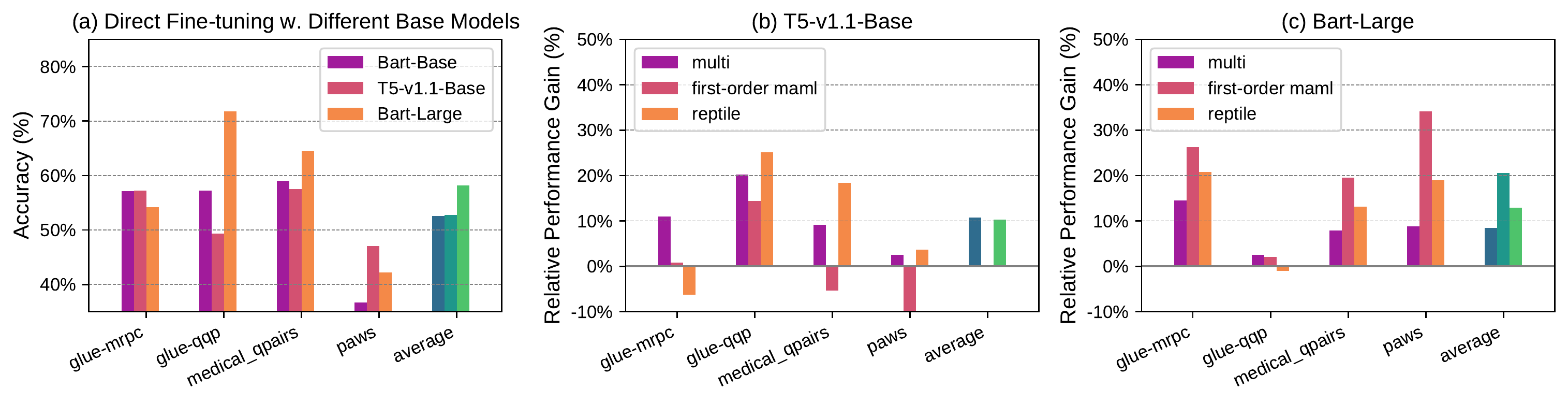}
    \caption{Extending upstream learning to larger pre-trained text-to-text models. (a) Absolute performance with direct fine-tuning with different pre-trained models. (b-c) Relative performance gain using upstream learning.}
    \label{fig:larger}
\end{figure*}

We have been mainly using BART-Base (139M parameters) as the main network, while it is possible to further push the limits of few-shot learning by using scaling up to larger models or using different model architectures. Previous work has shown that scaling up model size leads to better performance \cite{2020t5, brown2020language}. Moreover, since meta-learning algorithms are naturally unstable, it is important to verify whether they function as expected with larger models. In Q5, we experiment with T5-v1.1-Base (248M)\footnote{T5-Base was trained on a mixture of downstream tasks during its pre-training; such practice strays from the purpose of our study. Therefore, we use T5-v1.1-Base model, which is trained with the C4 Corpus only.} and BART-Large (406M) model with Held-out-Para Partition to verify these assumptions. We only consider first-order methods, as second-order optimization with these larger models is impossible with our available computation.

Our results are plotted in Fig.~\ref{fig:larger}. In Fig.~\ref{fig:larger}(a) we compare the few-shot performance of direct fine-tuning on these three pre-trained models. On average, few-shot performance grows with models size, with a few exceptions such as QQP+T5-v1.1-Base and MRPC+Bart-Large. In Fig.~\ref{fig:larger}(b-c) we plot the effect brought by upstream learning method for larger models. Except for FoMAML+T5-v1.1-Base\footnote{We observe instability in training loss during FoMAML training for T5-v1.1-Base.}, upstream learning methods consistently improves few-shot performance on $\mathcal{T}_{test}$, which verifies that upstream learning methods we use are model-agnostic, and can be applied to larger models to further improve few-shot performance.

\finding{Q6. Can we use pattern-exploiting training to replace direct fine-tuning to achieve even better performance?} 
Pattern-exploiting training (PET) is a novel method that formulate a target task into cloze-style questions \cite{schick2020exploiting,schick2020small,Gao2020MakingPL}. This approach narrows the gap between the masked language modeling objective during pre-training and downstream task fine-tuning, and therefore leads to more efficient transfer. PET is demonstrated to be effective with encoder models (e.g., RoBERTa), however, whether it is applicable to text-to-text models with auto-regressive decoders is underexplored to the best of our knowledge. In Q6, we study whether applying PET-style methods to text-to-text models is feasible, and whether combining the two methods further pushes the few-shot performance.

To align with the experiment settings in \cite{schick2020exploiting,schick2020small,Gao2020MakingPL}, we introduce a new task partition ``Held-out-GLUE'', which uses non-GLUE classification tasks as $\mathcal{T}_{train}$, and GLUE tasks as $\mathcal{T}_{test}$. We use the top 3 patterns in \cite{Gao2020MakingPL} for each GLUE task, and use the ensemble of the three models to produce the final prediction.

Since pattern-exploiting training is originally designed for encoder models (e.g., BERT/RoBERTa), we first tried two of its variants that adapts it to our auto-regressive transformer models. The first variant generates complete sentence, e.g., generate ``The movie is great. A wonderful piece'' from ``The movie is great. A <mask> piece'' for sentiment classification. The second variant generates only the word ``wonderful'', from ``The movie is great. A <mask> piece''. Though the first variant is more similar to the denoising pre-training objective of BART, we find the second variant to have better performance.

\begin{figure*}
    \centering
    \includegraphics[width=0.7\textwidth]{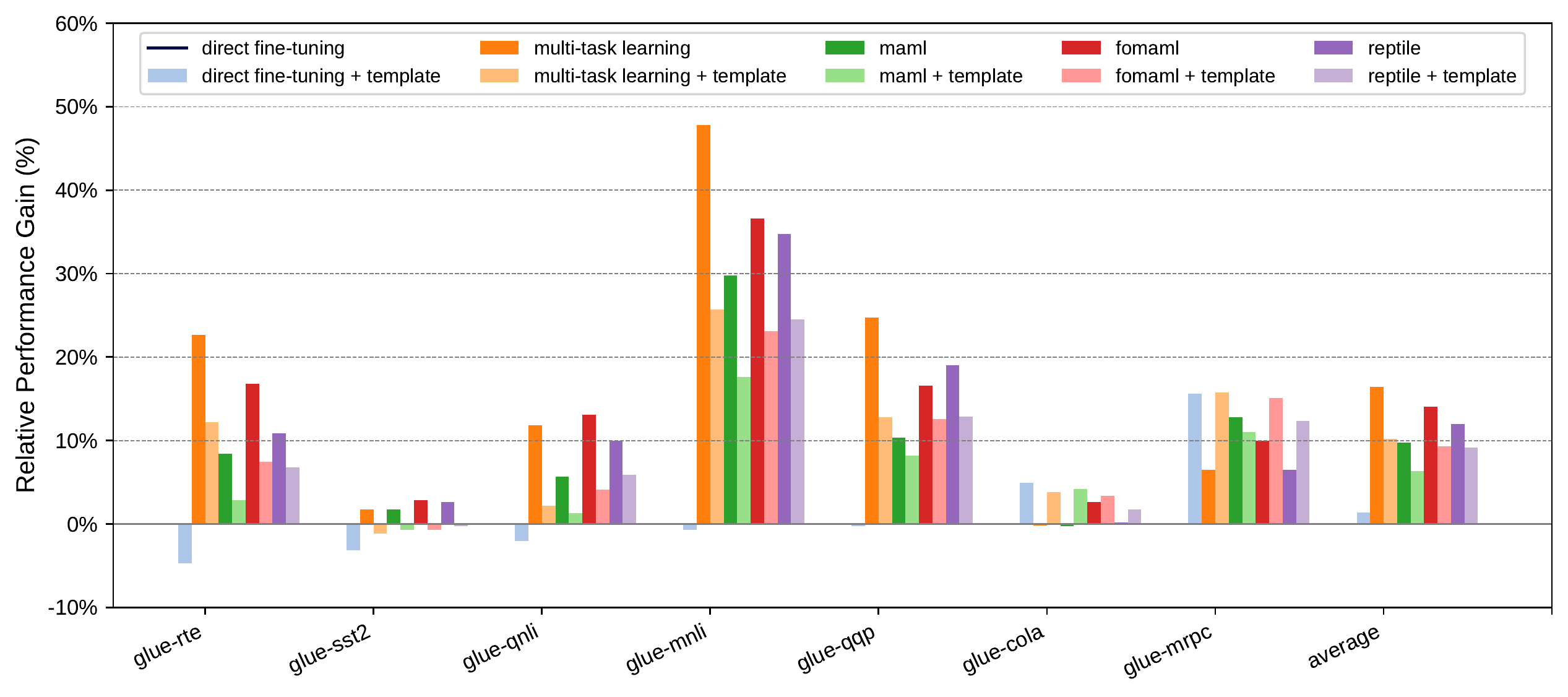}
    \caption{Combining upstream learning with pattern-exploiting training.}
    \label{fig:pet}
\end{figure*}

We then launch pattern-exploiting training using variant two with the original BART-Base models. We observe negative performance on average (leftmost blue bar in Fig.~\ref{fig:pet}). Performance is improved with CoLA and MRPC, but not with the remaining GLUE tasks. We further launch experiments with/without pattern-exploiting training, with our upstream learning checkpoints. Still pattern-exploiting training leads to deteriorated performance on average.

We stop further investigation since this is out of the scope of our study. Still we believe it is important to identify the reasons and develop pattern-exploiting methods for auto-regressive models.

\section{Reproducibility}
\paragraph{Implementation.} All our experiments are implemented with Huggingface Transformers\footnote{\url{https://github.com/huggingface/transformers}} \cite{wolf-etal-2020-transformers}. For higher-order optimization in the meta-learning approach optimization, we use \texttt{higher} library\footnote{\url{https://github.com/facebookresearch/higher}}. Our code has been uploaded in supplementary materials, and is also open-sourced at \url{https://github.com/INK-USC/CrossFit}.
\paragraph{Hyper-parameters.} We mainly follow the practice in \cite{Gao2020MakingPL}. During few-shot fine-tuning, we select the learning rate from $\{1e-5,2e-5,5e-5\}$, and the batch size from $\{2,4,8\}$, based on $D_{dev}$ performance. We set the total number of updates to be 1000, number of warmup updates to be 100. We evaluate the model on $D_{dev}$ every 100 steps.
\paragraph{Infrastructure and Runtime.} Upstream learning are done with one single Quadro RTX 8000 (48GB). Upstream learning jobs finishes within 3 hours on average.
Fine-tuning experiments are all done with one single GPU, with either NVIDIA Quadro GP100, NVIDIA Quadro RTX 8000, NVIDIA Quadro RTX 6000, NVIDIA GeForce RTX 1080 Ti, or NVIDIA GeForce RTX 2080 Ti, based on availability. Fine-tuning on one few-shot task (with hyperparmeter tuning for all 5 random samples) takes approximately 4 hours on average.
\paragraph{Number of Parameters.} BART-Base model contains 139 million parameters. T5-v1.1-Base model contains 246 million parameters. BART-Large model contains 406 million parameters.

\end{document}